%% file: main.tex
\documentclass[10pt,a4paper]{article}
\usepackage{blindtext}
\usepackage{multicol}
\usepackage{multirow}
\usepackage[
  backend=biber,
  style=authoryear-comp,
  sorting=nyt
]{biblatex}
\usepackage[pdfencoding=auto,psdextra,hidelinks]{hyperref}
\usepackage{graphicx} 
\usepackage[margin=25mm]{geometry}
\usepackage[font=sf]{caption} 
\usepackage{subcaption}
\usepackage{adjustbox}
\usepackage{amsmath}
\usepackage{ragged2e}
\usepackage{amsfonts}
\usepackage{amssymb}
\usepackage{siunitx}
\usepackage{float}
\usepackage{verbatim}
\usepackage{makecell}
\usepackage{tabularx}
\usepackage{xcolor}
\usepackage{makecell}
\usepackage{marvosym}
\usepackage{hyperref} 
\DeclareDelimFormat{nameyeardelim}{\addcomma\space}
\newcommand{\mytype}{Master's Thesis}

\newcommand{\mycourse}{Applied Data Science}

\newcommand{\mytitle}{Prompt Engineering: How Prompt Vocabulary affects Domain Knowledge}
\newcommand{\myauthor}{Dimitri Schreiter}
\newcommand{\mydepartment}{Institute of Computer Science}
\newcommand{\mysubmissiondate}{1. October 2024}
\newcommand{\myfirstsupervisor}{Prof. Dr. Bela Gipp}
\newcommand{\mysecondsupervisor}{Dr. Terry Lima Ruas}

\title{Prompt Engineering: How Prompt Vocabulary affects Domain Knowledge}
\author{Dimitri Schreiter}
\date{October 2024}

\begin{document}
\begin{titlepage}
    \normalsize
    \begin{tabularx}{\textwidth}{lXr}
        \multirow{2}{*}{\includegraphics[width=6.5cm]{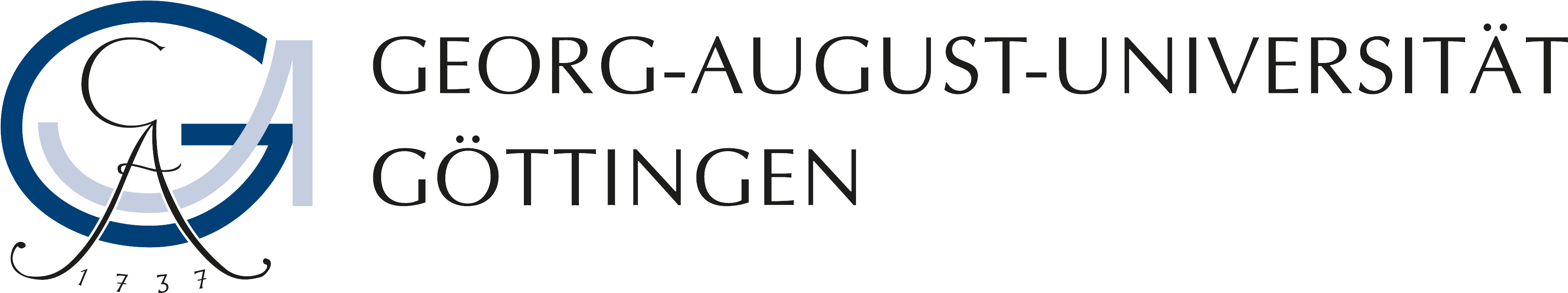}} 
        & & \\
    \end{tabularx}

    \large
    \centering

    \vspace{3cm}

    \textbf{\LARGE \mytype}\\

    submitted in partial fulfillment of the\\
    requirements for the course "\mycourse"

    \vspace{2cm}

    \textbf{\LARGE \mytitle}

    \vspace{2cm}

    \myauthor

    \vspace{2cm}

    \mydepartment

    \vspace{2cm}

    Master's Thesis\\
    of the Center for Computational Sciences\\
    at the Georg-August-Universität Göttingen

    \vspace{0.2cm}

    \mysubmissiondate

    \clearpage
    \thispagestyle{empty}
    \null
    \flushleft
    \normalsize

    \vspace{12cm}

    Georg-August-Universität Göttingen\\
    Institute of Computer Science\\[3ex]
    Goldschmidtstraße 7\\
    37077 Göttingen\\
    Germany\\[3ex]

    \begin{tabular}{@{}ll}
        \Telefon & +49 (551) 39-172000\\
        \fax & +49 (551) 39-14403\\
        \Letter & \href{mailto:office@informatik.uni-goettingen.de}{office@informatik.uni-goettingen.de}\\
        \Mundus & \url{www.informatik.uni-goettingen.de}\\
    \end{tabular}

    \vspace{1.0cm}

    \begin{tabular}{@{}ll}
        First Supervisor: & \myfirstsupervisor\\
        Second Supervisor:& \mysecondsupervisor\\
    \end{tabular}

    \clearpage
\end{titlepage}
\include{statement}
\tableofcontents
\newpage

\begin{multicols}{2}
\justifying
\begin{abstract}
    \small
    Prompt engineering has emerged as a critical component in optimizing large language models (LLMs) for domain-specific tasks. However, the role of prompt specificity, especially in domains like STEM (physics, chemistry, biology, computer science and mathematics), medicine, and law, remains underexplored. This thesis addresses the problem of whether increasing the specificity of vocabulary in prompts improves LLM performance in domain-specific question-answering and reasoning tasks. We developed a synonymization framework to systematically substitute nouns, verbs, and adjectives with varying specificity levels, measuring the impact on four LLMs: Llama-3.1-70B-Instruct, Granite-13B-Instruct-V2, Flan-T5-XL, and Mistral-Large 2, across datasets in STEM, law, and medicine. Our results reveal that while generally increasing the specificity of prompts does not have a significant impact, there appears to be a specificity range, across all considered models, where the LLM performs the best.  Identifying this optimal specificity range offers a key insight for prompt design, suggesting that manipulating prompts within this range could maximize LLM performance and lead to more efficient applications in specialized domains.
    \normalsize
\end{abstract}
\section{Introduction}
The rapid advancements in large language models have significantly expanded their applicability across various natural language processing (NLP) tasks. From zero-shot reasoning to few-shot learning, these models have demonstrated remarkable capabilities without requiring fine-tuning on specific datasets, as demonstrated by models such as GPT-3, T5, and Llama (\cite{brownLanguageModelsAre2020a, ouyangTrainingLanguageModels2022, touvron2023llamaopenefficientfoundation}). This versatility has enabled the deployment of LLMs in specialized fields such as STEM, medicine, and law, where accurate domain-specific responses are crucial. However, the growing reliance on LLMs has also underscored the importance of prompt engineering, the practice of crafting input prompts to elicit optimal performance from pretrained models. Unlike traditional fine-tuning approaches, which adjust model parameters based on specific datasets (\cite{liuPretrainPromptPredict2023a}), prompt engineering focuses on the design of the prompt itself to enhance model outputs without modifying model parameters (\cite{white2023promptpatterncatalogenhance, chen2024unleashingpotentialpromptengineering}). This approach is computationally efficient and scalable, particularly for large-scale applications.\\
A core challenge in prompt engineering is ensuring that the vocabulary and structure of the prompts align well with the model’s understanding of the task (\cite{zhengPromptLearningStructured2023, leidingerLanguagePromptingWhat2023}). One important but underexplored aspect of this is prompt specificity.\\ Specificity is a fundamental aspect of effective communication, especially within scientific and technical domains where precision is paramount (\cite{Ang2018Specificity}). The use of specific language reduces ambiguity, enhances clarity, and ensures that complex concepts are accurately conveyed and understood. In disciplines such as medicine, engineering, and law, selecting between words of varying specificity can impact interpretations and outcomes, making specificity a topic of particular interest. For example, in medicine, using the general term \textit{infection} versus the more specific synonym \textit{sepsis} could have critical implications. While \textit{infection} refers to the invasion and multiplication of microorganisms in the body, \textit{sepsis} is a specific, life-threatening response to infection that can lead to tissue damage and organ failure. Mislabeling sepsis as a general infection may delay necessary aggressive treatments, posing serious health risks to patients.\\
In engineering, referring to a \textit{metal} versus specifying \textit{titanium} could affect material selection and performance. \textit{Metal} is a broad category, whereas \textit{titanium} is a specific metal known for its high strength-to-weight ratio and corrosion resistance. Using the general term may result in inappropriate material choices, leading to design failures or safety hazards. Similarly, in law, the term \textit{crime} is general, whereas \textit{embezzlement} is a specific type of financial crime involving the unlawful taking of funds by someone in a position of trust. Confusing these terms could affect legal interpretations and sentencing; misclassifying \textit{embezzlement} as a general \textit{crime} may overlook the specific legal elements required for prosecution. Given these examples that underscore the importance of specificity in specialized fields, it leads us to the question whether this significance of specificity translates into the realm of LLMs, particularly in prompt engineering. Based on the previously mentioned benefits of using more specific vocabulary, we ask:

\textit{Does increasing the specificity of vocabulary in prompts enhance the performance of LLMs in generating responses in domain-specific question-answering and reasoning tasks?}\\ 

As LLMs are increasingly deployed in areas such as STEM, medicine and law, where precise communication is important, understanding the impact of prompt specificity on model outputs seems increasingly reasonable. This inquiry forms the basis of our study, motivating us to explore how varying the specificity of words within prompts affects the ability of LLMs to comprehend and accurately process domain-specific information.  By investigating this relationship, we aim to determine whether the benefits of specificity in human communication extend to interactions with LLMs, contributing to more effective and reliable applications of these models in specialized domains.\\
In this thesis, we examine the impact of prompt specificity on LLM performance in domain-specific question-answering and reasoning tasks. Specifically, we focus on three major parts of speech (nouns, verbs, and adjectives), and analyze how varying the specificity of these words in prompts influences the models’ ability to accurately answer questions. We calculate specificity scores for nouns and verbs, utilizing the lexical database WordNet (\cite{fellbaum1998wordnet}), and introduce a novel equation to quantify the specificity of adjectives. Our study evaluates four LLMs, Llama-3.1-70B-Instruct, Granite-13B-Instruct-V2, Flan-T5-XL, and Mistral-Large 2, across three datasets (MMLU, GPQA, and GSM8K), focusing on STEM, medicine, and law domains.\\
To summarize our contributions:
\begin{itemize}
    \item We introduce a method to systematically substitute nouns, verbs and adjectives through synonyms with different specificities.
    \item We find prompt specificity ranges for different models, where the LLM yields the best results in question-answering and reasoning tasks across STEM, law and medicine domains.
    \item We demonstrate that generally increasing prompt specificity, exceeding the optimal ranges of prompt specificity in the STEM, medicine, and law domains, has minimal impact on LLM performance for nouns in question-answering and reasoning tasks, but results in a significantly negative effect for verbs in reasoning NLP tasks.
    \item We introduce a method for calculating the specificity of adjectives, marking the first step towards ranking and quantifying adjectives, though further validation is necessary to confirm its significance.
\end{itemize}

\section{Related Work}
With the recent advancements in LLMs, their applicability to a wide range of NLP tasks has simultaneously expanded (\cite{brownLanguageModelsAre2020a}). These models exhibit complex capabilities, including zero-shot problem-solving (\cite{kojima2023large}), few-shot learning (\cite{cobbeTrainingVerifiersSolve2021}), instruction following (\cite{ouyangTrainingLanguageModels2022}), and incorporation of domain knowledge (\cite{zhengPromptLearningStructured2023, velasquez-henaoPromptEngineeringMethodology2023}). To harness these abilities, various prompt engineering techniques have emerged, aiming to interact with LLMs and significantly enhance performance across diverse NLP tasks (\cite{liuPretrainPromptPredict2023a, schickExploitingClozeQuestions2021}). Unlike fine-tuning methods that adjust model parameters by retraining with domain-specific labeled data, prompt engineering focuses on optimizing the input prompts to elicit the best possible outputs from a pretrained LLM without altering its parameters. Fine-tuning can be computationally resource-intensive, especially at scale, whereas prompt engineering typically requires only a few contextual examples for in-context learning  and minimal computational resources (\cite{dong2024surveyincontextlearning, liu2024understanding}).\\
Prompts, the core components of prompt engineering, can be categorized into continuous and discrete prompts. Continuous methods involve automated optimization of prompts, usually with a masked language model, by dynamically adjusting the prompt to ongoing interactions or contexts by rephrasing the baseline prompt or modifying factual details (\cite{shinAutoPromptElicitingKnowledge2020, ju-etal-2023-continuous, xia2024aprompt4em}). In contrast, discrete prompts are manually crafted, offering greater flexibility as they can be specifically tailored for distinct domains and use cases where the context remains relatively stable (\cite{schickExploitingClozeQuestions2021, reynoldsPromptProgrammingLarge2021}). In this work, we will focus on discrete prompts due to their differentiable properties and interpretability.\\
Several collections of discrete prompt templates exist, such as \textsc{PromptSource} (\cite{bachPromptSourceIntegratedDevelopment2022}), \textsc{Sup-NatInst} (\cite{wangSuperNaturalInstructionsGeneralizationDeclarative2022c}), and \textsc{BIG-BENCH} (\cite{srivastava2023imitation}), which are community-contributed resources designated for various NLP tasks. These collections facilitate performance quantification across multiple benchmarks and offer users convenient access to state-of-the-art prompts for specific NLP tasks.\\
While prompt engineering provides flexible means of interacting with LLMs by refining input prompts, the opaque nature of transformer-based learning can lead to unintended consequences, such as hallucinations (\cite{xu2024hallucinationinevitableinnatelimitation}). Consequently, carefully crafting prompts is critical to ensuring accurate and reliable outputs (\cite{rawteExploringRelationshipLLM2023}). Recent work by \cite{wahle2024paraphrasetypeselicitprompt} has shown that paraphrasing prompts significantly enhances LLM performance across various tasks, including sentiment analysis, question answering, and summarization. Their systematic evaluation of linguistic features revealed that certain elements, such as morphology, exert a stronger influence on model performance than others, like syntax, depending on the task.
Additionally, \cite{wangPromptEngineeringConsistency2024} highlighted the inconsistency of results generated by different prompts for the same task across multiple models in the medical domain, underscoring the need to align prompts with domain-specific knowledge. \cite{sorensenInformationtheoreticApproachPrompt2022} demonstrated that optimizing mutual information between the prompt and model output using unsupervised techniques yields high accuracy in various NLP tasks. Similarly, \cite{luHowArePrompts2023} identified a strong negative correlation between prompt sensitivity and LLM performance, proposing sensitivity-aware decoding to improve outcomes for prompts lacking sufficient contextual information. Focusing on linguistic dimensions such as modality, tense, and synonyms, \cite{leidingerLanguagePromptingWhat2023} further demonstrated that LLM performance is highly sensitive to both semantic and syntactic structures, and that transferring prompts across datasets and models often results in suboptimal performance.\\
The sensitivity of consistency and reliability concerning semantic structure has been thoroughly examined in current research. However, the influence of prompt specificity on model performance remains underexplored. \cite{zhengPromptLearningStructured2023} demonstrated that incorporating domain-specific vocabulary into prompts can substantially improve the performance of pretrained LLMs in both open and biomedical domains by iteratively modifying prompts in the word sense prediction (WSP) dataset through phrase additions or rephrasing. However, their study did not perform a stepwise analysis of individual word specificity or overall prompt specificity. Such a systematic examination could offer valuable insights into how specific word choices and prompt construction impact LLM behavior.\\
Our research builds upon the findings of \cite{zhengPromptLearningStructured2023} and \cite{leidingerLanguagePromptingWhat2023} by conducting a more detailed analysis of prompt specificity and its impact on LLM performance on question answering and reasoning tasks. While \cite{zhengPromptLearningStructured2023} demonstrated that incorporating domain-specific vocabulary into prompts could improve LLM performance in both open and biomedical domains, their study did not undertake a stepwise analysis of individual word specificity or overall prompt specificity. In contrast, our work systematically analyzes how individual word choices and prompt constructions influence LLM behavior across question answering and reasoning tasks, with a more diverse set of domains (STEM, law and medicine).
Similarly, \cite{leidingerLanguagePromptingWhat2023} observed that replacing words with non-standard synonyms can improve performance but did not provide a comprehensive breakdown of how specificity levels affect this outcome. Our study extends their findings by categorizing synonyms into distinct specificity levels (low, intermediate, high) and examining whether their observation holds true across different degrees of specificity. In this study, we address the limitations of previous studies by incorporating the dimension of specificity into prompt design and conducting a more nuanced investigation of synonym substitution with varying specificities and its effects on LLM performance. Further, this thesis aims to offer a deeper understanding of the interplay between prompt specificity and model behavior, extending across a diverse set of tasks and domains.

\section{Methodology}
In this thesis, we focus on the impact of prompt specificity on the performance of LLMs. Therefore, we create multiple variations of instructions from the datasets MMLU, GSM8K and GPQA (Section \ref{sec:data}), with our specificity-based synonymization framework (Section \ref{sec:framework}). This framework determines all parts of speech with the respective semantic sense in the provided context, retrieves the synonyms for each sense, calculates the specificity scores (Section \ref{sec:specificity}) by utilizing lexical database structures (Section \ref{sec:lex}), categorizes the synonyms based on the specificity and finally synonymizes the instruction with three replacement ratios (33\%, 67\% and 100\%) with respect to the replaceable word count. For each prompt variation, we use the models Granite-13B-Instruct-V2, Flan-T5-XL, Llama-3.1-70B-Instruct and Mistral-Large 2 (Section \ref{sec:models}) to generate the output for question-answering and reasoning tasks.
\clearpage
\end{multicols}
\begin{figure}[ht!]
\definecolor{lightred}{HTML}{fc5656}
\definecolor{lightblue}{HTML}{6eb2fa}
\definecolor{green}{HTML}{4c8736}

    \centering
    \includegraphics[width=0.90\textwidth]{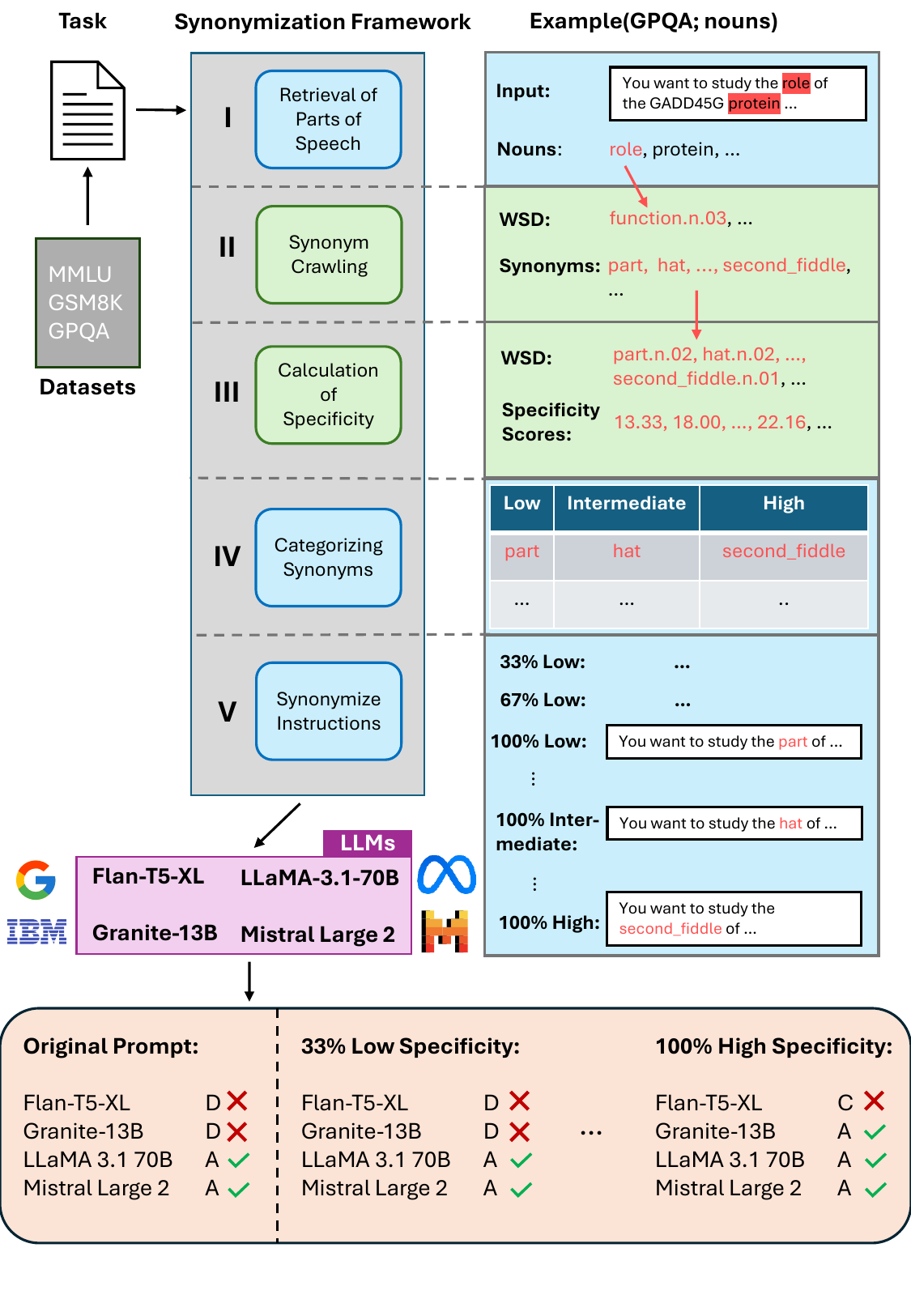}
    \caption{\textbf{Specificity-based Synonymization Framework.} Representation of the specificity-based synonymization framework used to synonymize the prompt instructions with varying specificities of all datasets. The preprocessing includes five key steps starting with the retrieval of parts (I) of speech from the original instruction, crawling synonyms (II) and calculate the specificity scores (III) for all parts of speech (the \textcolor{green}{green} colored boxes include WSD in the algorithm), categorizing the synonyms (IV) into low, intermediate, high specificity and finally synonymize the original instructions (V) with three different replacement ratios (33\%, 67\%, 100\%). Additionally, there is a step-by-step example sampled from the GPQA dataset, synonymizing \textcolor{lightred}{nouns} with varying specificity synonyms.}
    \label{fig:flow}
\end{figure}
\clearpage
\begin{multicols}{2}
\begin{table}[H]
    \centering
    \adjustbox{max width=\linewidth}{
    \begin{tabular}{|c|c|c|c|c|}
    \hline
    & \multicolumn{1}{c|}{Original} & \multicolumn{3}{c|}{Processed Samples}\\ 
    \cline{3-5}
    Dataset & Samples & Nouns & Verbs &Adjectives\\
    \hline
    GPQA & 250 & 181 & 154 &23 \\
    GSM8k & 250 & 174 &  159 &4 \\
    MMMLU & 1790 & 1376 &985  &145 \\
    \hline
    \end{tabular}}
    \caption{\textbf{Overview of data.} This table shows the number of samples before and after the preprocessing steps, indicating that nouns and verbs are suitable and significantly enough represented in the data, while adjectives seem underrepresented.}
    \label{tab:data}
\end{table}
\subsection{Synonymization Framework}
\label{sec:framework}
The specificity-based synonymization framework, shown in Fig. \ref{fig:flow}, consists out of five distinct steps, starting from the retrieval of the considered parts of speech (I), crawling of synonyms (II), calculating of specificity scores (III), categorizing the synonyms according to the specificity scores (IV) and the synonymization of the instructions (V). In the following, we will explain each step and additionally explain our approach of calculating the overall prompt specificity, focusing on one part of speech at a time. Note, that we only use a fraction of the complete datasets, since the number of synonyms increases exponentially with the number of instructions and therefore increases the computational time significantly, although all effort was made to optimize the computational efficiency of the framework. For each dataset (MMLU with 14 subdatasets, GPQA and GSM8k) we sampled up to 250 instructions. The overview of the samples remained after the processing is displayed in Tab. \ref{tab:data}.\\
\textbf{I Part of Speech Retrieval.} From the original instruction prompt, we parse the sentence by word-tokenizing the instruction, tagging the lexical database position and filter for nouns, verbs and adjectives from each individual prompt\footnote{We used word tokenize and parts of speech tagging functions in the NLTK python package with the universal tagset (\cite{petrov2011universalpartofspeechtagset}): https://www.nltk.org/}. This algorithm yields us lists of all considered parts of speech that we pass to the next step.\\
\textbf{II Crawling synonyms.} For each tagged word, we apply Word Sense Disambiguation (WSD) (\cite{wahle2021incorporating}) to identify the most contextually appropriate sense. We then retrieve all associated hypernyms, lemmas, and hyponyms for the selected sense. This approach helps to consider semantically suitable synonyms, reducing the likelihood of confusing the LLM by selecting synonyms with incorrect senses.
 For instance, the noun \textit{cell}, which appears in a biological context and refers to \textit{the basic structural and functional unit of all organisms}, has six other senses in the lexical database:\\
\textit{cell as ...}
\begin{itemize}
    \item \textit{... any small compartment}
    \item \textit{... a device that delivers an electric current as the result of a chemical relation}
    \item \textit{... a small unit serving as part of or as the nucleus of a larger political movement}
    \item \textit{... a hand-held mobile radiotelephone for use in an area divided into small sections, each with its own short-range transmitter/receiver}
    \item \textit{... small room in which a monk or nun lives}
    \item \textit{... a room where a prisoner is kept}
\end{itemize}
Each of these senses would be connected to synonyms (e.g. \textit{cubicle} or \textit{jail cell}) which would not be fitting in our biological context, potentially changing the meaning of the sentence and therefore confuse the LLM. \newline
\textbf{III Calculating specificity scores.} The gathered synonyms are then passed to the specificity calculation algorithm, which quantifies the specificity for each synonym with a continuous number by utilizing the taxonomy structures of the lexical database for each part of speech (Section \ref{sec:specificity})\\
\textbf{IV Categorizing Synonyms.} Using the calculated specificity scores, we will select three synonyms from the set of all possible synonyms for each part of speech, categorizing them into low, intermediate, and high specificity based on the minimum, mean, and maximum values, respectively. This requires that each word must have \textit{at least three unique specificity scores}, as otherwise, it would be impossible to fill all three categories. The intermediate specificity is selected by calculating the minimum absolute difference $\min |s - \Bar{s}|$ between the mean specificity score $\Bar{s}$ of the set and the synonym's specificity score $s$, choosing the value closest to the mean of the set.\newline
\textbf{V Synonymize instructions.}
To generate synonymized instructions, we fractionally replace the parts of speech in the prompt instructions with their categorized synonyms at rates of 33\%, 67\%, and 100\% focusing on one type at a time. This introduces a second constraint: \textit{Each instruction must contain at least three different replaceable parts of speech}. The algorithm yields nine different synonymized instructions for each part of speech, in addition to the original prompt from the dataset. An example of 100\% replacement of nouns with synonyms of low, intermediate, and high specificity can be found in Appendix \ref{sec:appendixdata}, Tab. \ref{tab:examples}.\\
\end{multicols}
\begin{table}[ht!]
    \centering
    \adjustbox{max width=\textwidth}{
    \begin{tabular}{|c|c|c|c|c|c|c|}
    \hline
    Part of Speech & Hypernyms & Hyponyms & Average \#synsets (Hypernym) & Average \#synsets (Hyponym)& Stat & p-value\\
    \hline
    \hline
    Nouns& 7890 & 28021 & 3.55 & 2.41 & 135e9 & 1.46e-222 \\
    Verbs& 12646 & 27454 & 7.90 & 5.32 & 216e9 & 0.0\\
    \hline
    \end{tabular}
    }
    \caption{\textbf{Average Number of synsets evaluation.} Results of the assessment of average number of synsets for hyper- and hyponyms. A Mann-Whitney U test was performed to quantify whether the mean of synsets of hypernyms is significantly greater than the mean of synsets of hyponyms.}
    \label{tab:just}
\end{table}
\begin{multicols}{2}
\subsection{Lexical Database}
\label{sec:lex}
The lexical database that we use for this study is WordNet (WN), introduced by the Princeton University as an open source project in 1985. The database includes English nouns, verbs, adjectives and adverbs, which are structured through cognitive synonyms called \textit{synsets} (distinct semantic concepts) that are interconnected through static relations (\cite{10.1145/219717.219748}).\\
Noun synsets are structured hierarchically through super-subordinate relations (hypernymy and hyponymy). Each noun hierarchy starts with the ENTITY root synset and unfolds into an inverse tree. Synsets at the top of the tree are more general and become more specific moving down the tree. Non-leave nodes are considered types and represent common nouns, whereas leafs of the tree are instances that include personas, countries and geographic entities. Selecting one particular noun synset in the taxonomy, hypernyms represent all nodes that are between the root node and the chosen synset, while hyponyms represent all more specific synsets under the chosen synset (\cite{fellbaum1998wordnet}).\\
Verb synsets are organized similarly to noun synsets, forming hierarchies connected by various semantic relations, including not only hypernyms and hyponyms but also entailment, antonymy, and troponymy. Entailment describes a logical relationship between verbs where one implies the other, such as \textit{snore} implying \textit{sleep}. Antonymy refers to verbs representing opposite actions, for example, \textit{start} is the antonym of \textit{stop}. Troponymy, on the other hand, refers to the manner in which an action is performed, indicating a more specific way of carrying out the verb. For instance, \textit{jog} is a troponym of \textit{run} (\cite{fellbaum1998wordnet}).\\
In contrast, adjectives form a cluster-like structure composed of synonyms, antonyms and semantically similar relationships. Synonyms refer to adjectives that share the same conceptual meaning, such as \textit{happy} and \textit{joyful}, while semantically similar adjectives, e.g. \textit{large} and \textit{big} are linked through associative connections (\cite{fellbaum1998wordnet}).
\subsection{Specificity Measures}
\label{sec:specificity}
The specificity quantification for nouns, verbs and adjectives is based on the utilization of the corresponding WN taxonomy structures. For nouns and verbs, this taxonomy is hierarchically structured, allowing us to utilize this natural ordering of the lexical database to determine the specificity of the parts of speech. However, adjectives do not share this hierarchical structure with nouns and verbs in WN. Therefore, we use a novel adjective specificity measure to quantify their specificity. In the following, we will consider both equations and explain them in detail.\\
\textbf{Specificity Score for Nouns and Verbs.} Using the specificity score introduced by Bolognesi et al. (2020) (\cite{bolognesiAbstractionDecouplingConceptual2020}), we compute the specificity for noun and verb synonyms based on the following formula:
\begin{align}
    \mathrm{S_{\mathrm{noun/verb}}} = d + \log\left(\frac{1+n}{N}\right) - \log(l),
    \label{eq:specmes}
\end{align}
where \(d\) is the distance (number of nodes) between a word and the top root node, \(N\) represents the total number of nodes in the WN taxonomy restricted to a part of speech, \(n\) denotes the total number of direct and indirect hyponyms for a given word, and \(l\) is the number of synsets for the corresponding part of speech. Given that the taxonomy for nouns and verbs is hierarchically structured, with specificity increasing downwards, this equation aims to determine the relative position of a synset along its path in the respective WN taxonomy. The first term \(d\) counts the number of nodes from the root to the synset of interest. The second term  \(\log\left(\frac{1+n}{N}\right)\) represents the number of nodes directly and indirectly connected to the synset, normalized by the total number of nodes in the taxonomy for better comparability. By considering the nodes above and below, we can locate the synset within its respective path in the taxonomy. The term \(\log(l)\), which we added to the original equation, is based on the intuition that more specific words have less distinct senses. Our analysis of 940 nouns and 745 verbs shows that the average number of synsets for hypernyms is significantly higher than for hyponyms. The overview of the assessment with the corresponding results of the Mann-Whitney U test (\cite{doi:https://doi.org/10.1002/9780470479216.corpsy0524}) are 
\begin{figure}[H]
    \centering
   \includegraphics[width=\linewidth]{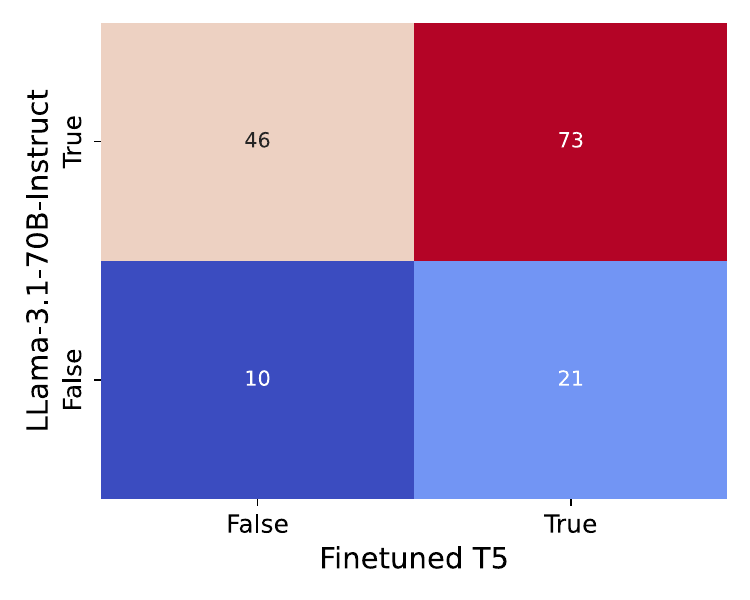}
    \caption{\textbf{Confusion Matrix of WSD Model Evaluation.} The performance agreement of Llama-3.1-70B-Instruct and finetuned T5 for WSD when predicting against the human evaluated ground truth.}
    \label{fig:conf}
\end{figure}
\begin{table}[H]
    \centering
    \begin{tabular}{|c|c|}
    \hline
         Model & Accuracy  \\
         \hline
    \hline
         Fine-Tuned T5& $0.63 \pm 0.05$\\
         Llama-3.1-70B-Instruct & $0.79 \pm 0.02$\\\hline
\end{tabular}
\caption{\textbf{WSD Model Evaluation.} Results of WSD performance assessment for fine-tuned T5 and Llama-3.1-70B-Instruct models for 50 mixed sampled (nouns and verbs) over three seeds.}
    \label{tab:wsd}
\end{table}
displayed in Tab. \ref{tab:just}. Given that hypernyms are inherently broader in meaning, while hyponyms convey greater specificity, this observation aligns with our intuition that words with fewer synsets tend to be more specific. Furthermore, we tackled a challenge where synonyms frequently belonged to the same synset as the target word, yielding identical specificity scores under the original measure. To address this, we integrated the total number of synsets associated with each synonym, effectively mitigating this issue.\\
\cite{bolognesiAbstractionDecouplingConceptual2020} used only the first sense of words for specificity calculations, as it represents the most common sense. However, this approach may not be ideal for our use case, as it cannot always ensure the correct sense for a given context is chosen, 
which could alter the semantic meaning of the sentence and might mislead the LLM. To address this, we apply WSD (\cite{wahle2021incorporating}) to ensure the synonym is accurately aligned with the context.\\
For this task, we selected the Llama-3.1-70B-Instruct model, which offers faster processing and significantly reduces computation time compared to the fine-tuned T5 model used by \cite{wahle2021incorporating}. To validate this model substitution, we benchmarked it against 
\begin{table}[H]
    \centering
    \adjustbox{max width = \linewidth}{
    \begin{tabular}{|c|c|c|c|c|}
\hline
\multirow{2}{*}{Parameter} & \multicolumn{4}{c|}{Adjectives} \\ \cline{2-5} 
 & good & phenomenal & small & tiny \\ \hline
ssw & 56 & 1 & 63 & 2 \\ \hline
s & 32 & 1 & 13 & 7 \\ \hline
a & 2 & 0 & 2 & 0 \\ \hline
l & 21 & 2 & 10 & 1 \\ \hline
\end{tabular}}
    \caption{\textbf{Comparison of Adjective Specificity Measure Parameters.} The number of words with similar meaning \(ssw\), number of direct synonyms \(s\), number of direct and indirect antonyms \(a\) and number of senses \(l\) of the adjectives \textit{good}, \textit{phenomenal}, \textit{small} and \textit{tiny}.}
    \label{tab:resexamples}
\end{table}
the fine-tuned T5 model using 50 mixed samples (nouns and verbs) on three different seeds. As demonstrated in Table \ref{tab:wsd}, Llama-3.1-70B-Instruct achieved higher accuracy (0.79 $\pm$ 0.02) compared to the fine-tuned T5 (0.63 $\pm$ 0.05).\\
We applied the McNemar test (\cite{dror-etal-2018-hitchhikers}) across three random seeds using the confusion matrix depicted in Fig. \ref{fig:conf} to assess whether the observed performance differences between the models are statistically significant. 
 The test returned a p-value of 0.003 with a test statistic of 21. Out of 67 cases where the models produced divergent results (46 + 21), the Llama-3.1-70B-Instruct model correctly identified the sense in 46 instances. The low p-value suggests that the performance disparity between the two models is statistically significant across the three runs. Given that Llama-3.1-70B-Instruct consistently outperforms the fine-tuned T5 model, it emerges as a viable alternative for the WSD process, offering both enhanced efficiency and accuracy\\
\textbf{Specificity Score for Adjectives.} As previously mentioned, adjectives do not share the same structural properties as nouns and verbs in WN, lacking a hierarchical arrangement. Therefore, we propose a new equation to measure adjective specificity:
\begin{equation}
    S_{\mathrm{adjectives}} = \frac{1}{\log(1+ssw+s+a+l)}
\end{equation}
where \(ssw\) represents the number of words with semantically similar relation to the considered adjective, \(s\) is the number of synonyms, \(a\) is the count of antonyms for the adjective and its synonyms, and \(l\) refers to the number of synsets for the adjective. The derivation of this equation is based on four underlying assumptions:
\begin{enumerate}
    \item The number of words with similar meaning \(ssw\) as the adjective is reverse proportional to its specificity \(S\sim\frac{1}{ssw} \)
    \item The number of direct synonyms \(s\) of an adjective is reverse proportional to its specificity \(S\sim\frac{1}{s} \)
    \item The number of direct and indirect antonyms \(a\) of an adjective is reverse proportional to its specificity \(S\sim\frac{1}{a} \)
    \item The number of senses \(l\) of an adjective is reverse proportional to its specificity \(S\sim\frac{1}{l} \)
\end{enumerate}
The underlying premise of these assumptions is that as the specificity of adjectives increases, their contextual meaning becomes more constrained. In other words, if an adjective can be substituted by many others (e.g., similar words, senses, direct synonyms, or even antonyms), it likely captures a broader range of features that those adjectives share. Consider the following examples: \textit{good} vs. \textit{phenomenal} and \textit{small} vs. \textit{tiny}. In both cases, the second adjective is more specific due to its narrower scope of applicability. For instance, while many things can be considered \textit{good}, not all good things are \textit{phenomenal}. Likewise, although everything \textit{tiny} is \textit{small}, not all \textit{small} things are \textit{tiny}. The results derived from parameter calculations based on these assumptions are summarized in Tab. \ref{tab:resexamples}.
The parameters for \textit{phenomenal} and \textit{tiny} are consistently smaller than those for \textit{good} and \textit{small}, which aligns with our assumptions. By incorporating all the assumed proportional relationships into an additive model for adjective specificity, we derive the following expression:
\begin{align}
    S \sim \frac{1}{ssw + s + a + l}.
    \label{eq:adjspec}
\end{align}

Each added term enhances the interpretability of the equation, as all parameters are treated independently. The additive model also reduces sensitivity to outliers and extreme values compared to the multiplicative model. To smooth the resulting specificity measure and make it more responsive to small changes, we apply a \( \log \) function to the denominator and add \( +1 \) to prevent division by zero, leading to our final equation \ref{eq:adjspec}.

To validate this approach, we used GPT-4o as an evaluator, comparing our systematic ranking of adjective specificity in various contexts with GPT-4o's rankings for a sample of 91 instructions. For each instruction, we filtered all adjectives, retrieved their respective synonyms, and iteratively created synonym pairs by incrementing the index by 1:
\begin{align*}
     [\text{syn}_1, \text{syn}_2, \text{syn}_3 ]&\rightarrow [\text{syn}_1, \text{syn}_2], \\
    &\ \ \ \ \ [\text{syn}_2, \text{syn}_3], \\
    &\ \ \ \ \ [\text{syn}_3, \text{syn}_1]
\end{align*}
These pairs, along with the original instruction, were then passed into a three-shot prompt template for GPT-4o, shown in Appendix \ref{sec:appendixa}, Fig. \ref{fig:prompts}. 
\begin{figure}[H]
    \centering
   \includegraphics[width=\linewidth]{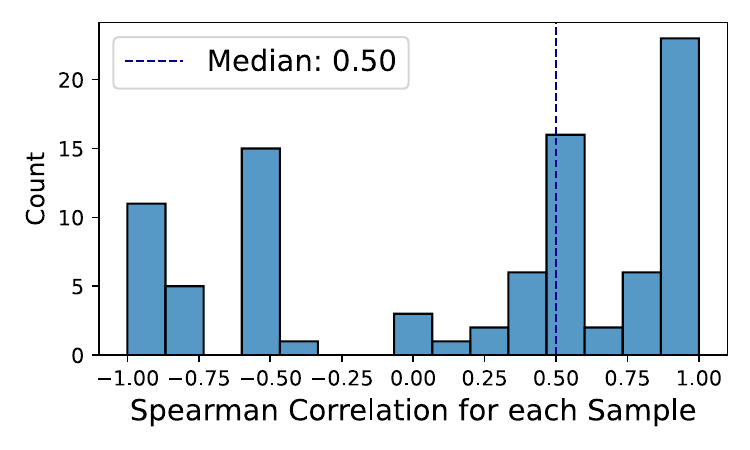}
    \caption{\textbf{Spearman-Correlations for Adjective Ranking.} The histogram displays the distribution of Spearman correlations from the LLM-as-a-judge experiment, which compares the model's qualitative ranking of adjective specificity with the calculated specificity score measure for 91 samples. The median Spearman correlation in this distribution is 0.50.}
    \label{fig:adjspec}
\end{figure}
The model generated an ordered list where the first element represents the adjective with greater specificity, given the context of the instruction. By applying this process to all adjective pairs, we used transitivity
\begin{align*}
    \left( A > B \land B > C \right) \implies A > C
\end{align*}
to combine all sub-rankings into a single, complete ranking of adjectives for each instruction.\\
We then correlated these GPT-4o adjective rankings with those derived from our adjective specificity model. The resulting Spearman correlations are shown in Fig. \ref{fig:adjspec}.
The median Spearman correlation coefficient of 0.50 indicates a moderate alignment between the equation-based and LLM-based rankings of adjective specificity, demonstrating that the methods are generally consistent. Notably, in 23 cases, a perfect positive correlation of 1.0 was observed, where both methods produced identical rankings, suggesting that the equation-based approach is highly effective in many instances. While 32 out of the 91 samples exhibited negative correlations, this divergence underscores the potential for refining the equation-based model to further improve alignment with LLM judgments, but the overall results reflect a solid foundation for its use in ranking adjective specificity.\\
\textbf{Calculating Prompt Specificity.}
\label{sec:promptspec}
The previous calculations focused solely on individual parts of speech and their corresponding synonym specificities, without considering the full composition of parts of speech within the prompt. To provide a broader perspective, we also evaluate overall changes in prompt specificity by converting the ordinal categories (33\% Low Specificity,..., 100\% High Specificity) into continuous specificity 
\begin{figure}[H]
    \centering
   \includegraphics[width = \linewidth]{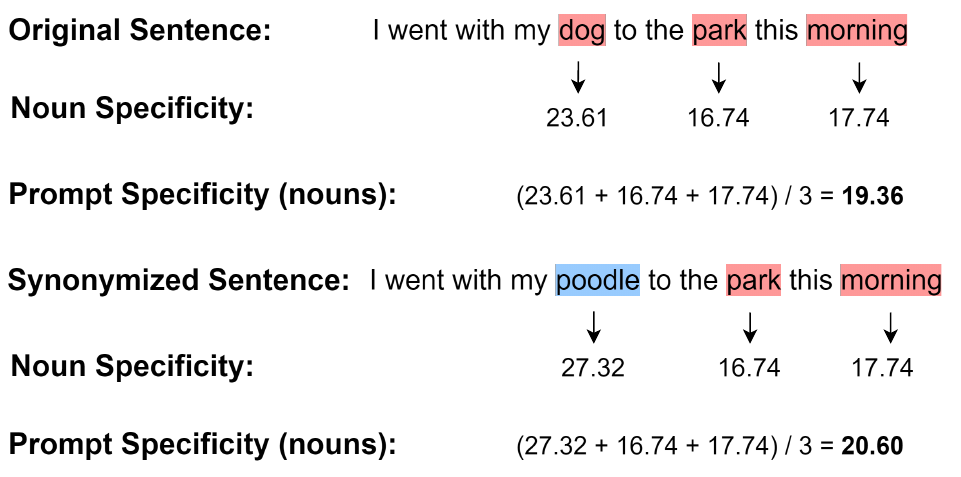}
    \caption{\textbf{Example for Prompt Specificity Calculation.} This example schematically illustrates the calculation of the prompt specificity, by aggregating the specificities of one part of speech (nouns in this case) and calculating the average that we call prompt specificity. Additionally, it shows the prompt specificity change from 19.36 to 20.60 after we substitute the noun \textit{dog} with the more specific synonym \textit{poodle}.}
    \label{fig:promptspecexample}
\end{figure}
scores. This is achieved by aggregating the individual specificity scores for each part of speech (nouns, verbs, and adjectives) within each instruction and calculating their average. As a result, each prompt is assigned a continuous specificity score for the considered part of speech, which varies across specificity levels and replacement levels
A simple example that illustrates this method can be seen in Fig. \ref{fig:promptspecexample}.
\subsection{Data}
\label{sec:data}
The datasets utilized in this study span diverse domains, including STEM, law, and medicine, as well as varying levels of expertise, ranging from high school and undergraduate to PhD-level. These datasets also encompass different task formats, such as multiple-choice questions (MMLU, GPQA) and reasoning-based assessments (GSM8K), ensuring a robust foundation for our evaluation. A detailed summary of all datasets including examples is presented in Appendix \ref{sec:appendixdata}.

\textbf{MMLU.} The \textit{Massive Multitask Language Understanding} dataset is an evaluation benchmark designed to assess the performance of models across a wide range of subjects and expertise levels. It contains approximately 57 tasks, each with up to 500 samples, accompanied by corresponding ground truths. These tasks are divided into various categories, including STEM, law and medicine in different expertise levels. The tasks include multiple-choice questions that span both general knowledge and highly specialized domains. The diversity of the MMLU dataset allows us to evaluate how well models generalize across a variety of topics, expertise levels, and formats, ensuring that the findings of this study are applicable to a broad range of real-world applications (\cite{hendryckstest2021}).

\textbf{GPQA.} The \textit{Google-Proof Question Answering} dataset consists of 448 expertly curated multiple-choice questions, with a focus on biology, physics, and chemistry. Each question is accompanied by a domain expert-evaluated ground truth, ensuring high-quality, reliable answers. The dataset is designed to test models at a PhD level of expertise, making it particularly challenging for assessing the performance of LLMs in specialized fields that require in-depth knowledge and understanding. This dataset is particularly useful for testing the robustness of models in handling precise, domain-specific questions that require not only basic factual knowledge but also a deep comprehension of scientific principles and reasoning (\cite{rein2023gpqagraduatelevelgoogleproofqa}.

\textbf{GSM8K.} The \textit{Grade School Math 8K} dataset is a carefully curated collection of 8,500 human-written math problems, each designed to reflect the complexity of grade-school-level arithmetic. The problems typically require two to eight steps to solve, involving sequential execution of arithmetic operations such as addition, subtraction, multiplication, and division. The dataset tests the model’s ability to reason through multistep problems, handle intermediate results, and execute operations in the correct order, which are crucial skills for successful mathematical problem-solving. The expertise level targeted by this dataset corresponds to middle school students, and solving these problems provides a clear benchmark for evaluating models' capacity for stepwise reasoning. For this study, GSM8K will serve as a key resource for assessing the models’ abilities to handle arithmetic reasoning tasks with chain-of-though prompting, a critical area in understanding the reasoning capabilities of LLMs in structured problem-solving contexts (\cite{cobbe2021gsm8k}).
\subsection{Models}
\label{sec:models}
The following models were selected due to the diverse architectures and sizes (3B, 13B, 70B, 123B) they present, while offering the capabilities of handling complex natural language processing tasks across specialized domains such as question-answering (zero-shot and few-shot), allowing a comprehensive analysis of prompt specificity. For each dataset, we use different prompt templates from the data sources to perform the question-answering and reasoning tasks (\cite{hendryckstest2021, rein2023gpqagraduatelevelgoogleproofqa, cobbe2021gsm8k}). We call the models using a 
\end{multicols}
\clearpage
\begin{figure}[ht!]  
    \centering
    \begin{subfigure}[b]{0.45\textwidth}
       \includegraphics[width=\textwidth]{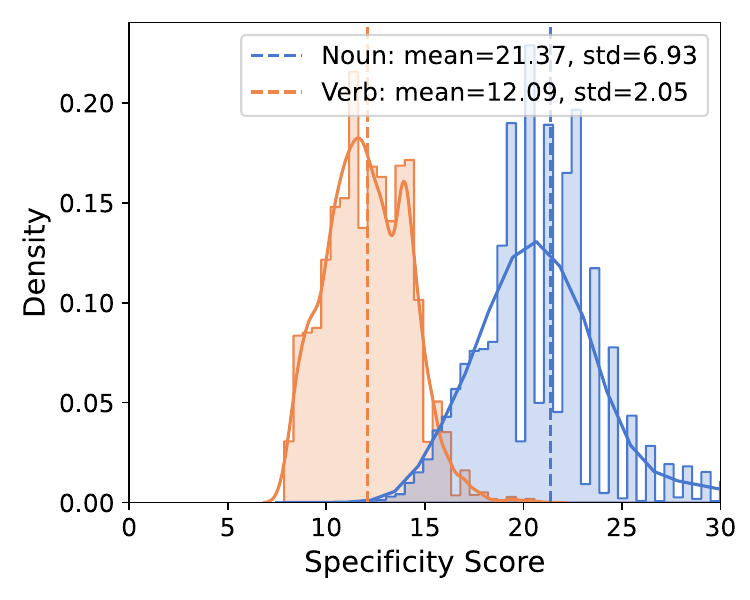} 
        \caption{\textbf{Nouns and Verbs.}} 
        \label{fig:spec_distr_nv}
    \end{subfigure}
    \hspace{0.05\textwidth}
    \begin{subfigure}[b]{0.45\textwidth} 
       \includegraphics[width=\textwidth]{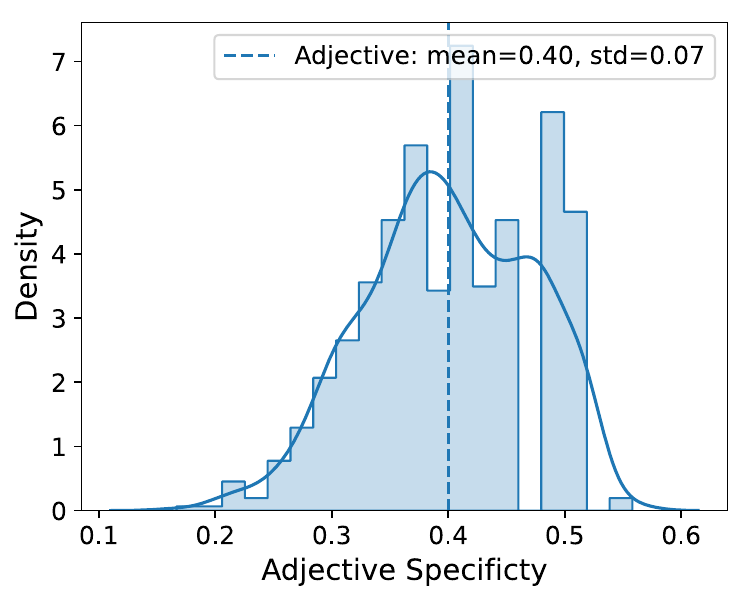} 
        \caption{\textbf{Adjectives.}}
        \label{fig:spec_distr_a}
    \end{subfigure}
    \caption{\textbf{Specificity Score Distribution for nouns, verbs and adjectives.} The histograms show the distribution
of specificity scores for the respective part of speech. The mean specificity
score for nouns is $\mu_{\text{nouns}} = 21.37$, for verbs $\mu_{\text{verbs}} = 12.09$ and for adjectives $\mu_\text{adjectives} = 0.40$}  
    \label{fig:distr}  
\end{figure}
\begin{multicols}{2}
temperature of zero to select only tokens with the highest probability, since question-answering and reasoning task require a high level of precision and determinism. Further, we set a threshold to the maximum output token size to 500, because during initial experimental runs, we observed that sometimes models repeat phrases until there is a timeout, stopping the process. To mitigate this issue, we stop the output generation process after the 500 token threshold is reached. For reproducibility, we set a random seed.
In all our experiments, we use a zero-shot prompt-templates for the MMLU and GPQA datasets and a zero-shot Chain-of-Thought (CoT) prompt template for the GSM8K dataset. An overview of the prompt templates and the parameters are displayed in Appendix \ref{sec:appendixa} Fig. \ref{fig:prompts} and Appendix \ref{sec:appendixb} Tab. \ref{tab:params}, respectively.\\
\textbf{Flan-T5-XL.}  Flan-T5-XL (3B) is a 3 billion parameter model that belongs to the Flan-T5 family. It is based on Google's T5 architecture, which uses an encoder-decoder model, and is pretrained on a mixture of supervised and unsupervised tasks that have been reformulated into a text-to-text format. The model is fine-tuned on the Fine-tuned LAnguage Net (FLAN) dataset, which incorporates instruction-based tuning to enhance its capabilities in zero-shot and few-shot learning scenarios (\cite{https://doi.org/10.48550/arxiv.2210.11416}).\\
\textbf{Granite-13B-Instruct-V2.} Granite-13B-Instruct-V2 is a general-purpose decoder-only model developed by IBM. This 13-billion-parameter model is optimized for a variety of NLP tasks through instruction tuning. It builds on the Granite-13B-V2 base, which was pretrained on 2.5 trillion tokens sourced from IBM's Data Pile. Granite-13B-Instruct-V2 has been fine-tuned to handle a broad range of tasks including text generation, comprehension, and question-answering, making it well-suited for zero-shot and few-shot learning paradigms (\cite{ibm2024granite}).\\

\begin{table}[H]
    \centering
    \begin{tabular}{|c|c|c|}
    \hline
        Part of Speech & Statistic & p-value \\
        \hline
         Nouns & 0.233&0.0\\
         Verbs &0.068&3.15e-34\\
         Adjectives &0.087&1.08e-5\\
         \hline
    \end{tabular}    
    \caption{\textbf{Results of Kolmogorov-Smirnov Test.} The table displays the results for the Kolmogorov-Smirnov test, which was performed on the specificity score distribution for nouns, verbs and adjectives. Since all p-values are smaller than 0.05, we cannot reject the null hypothesis and therefore the distributions cannot be considered as normal distributions.}
    \label{tab:kstest}
\end{table}
\textbf{Mistral-Large 2.} Mistral Large 2 is a 123 billion parameter model designed for NLP tasks with a focus on code generation, reasoning, and multilingual support. The model is fine-tuned to minimize hallucinations and follows instructions more accurately, making it effective in complex, long-context applications (\cite{mistral2024large2}).\\
\textbf{Llama-3.1-70B-Instruct.} Llama-3.1-70B-Instruct is part of Meta’s Llama series, known for their high performance in a wide variety of NLP tasks.
With 70 billion parameters, this model leverages its size to capture a wide range of linguistic and semantic nuances. The Llama series has been pretrained on vast amounts of data and subsequently fine-tuned through instruction-based learning and preference tuning, which enhances its ability to follow detailed and specific prompts (\cite{dubey2024llama3herdmodels}).\\ 
\end{multicols}
\begin{figure}[ht!]  
    \centering
    \begin{subfigure}[b]{0.45\textwidth}
       \includegraphics[width=\textwidth]{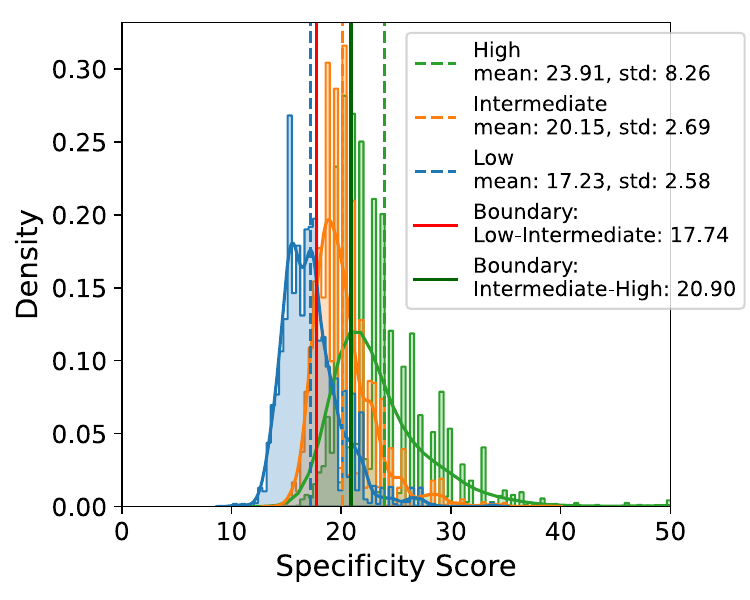} 
        \caption{\textbf{Nouns.}}
        \label{fig:boundn}
    \end{subfigure}
    \hspace{0.05\textwidth}
    \begin{subfigure}[b]{0.45\textwidth} 
       \includegraphics[width=\textwidth]{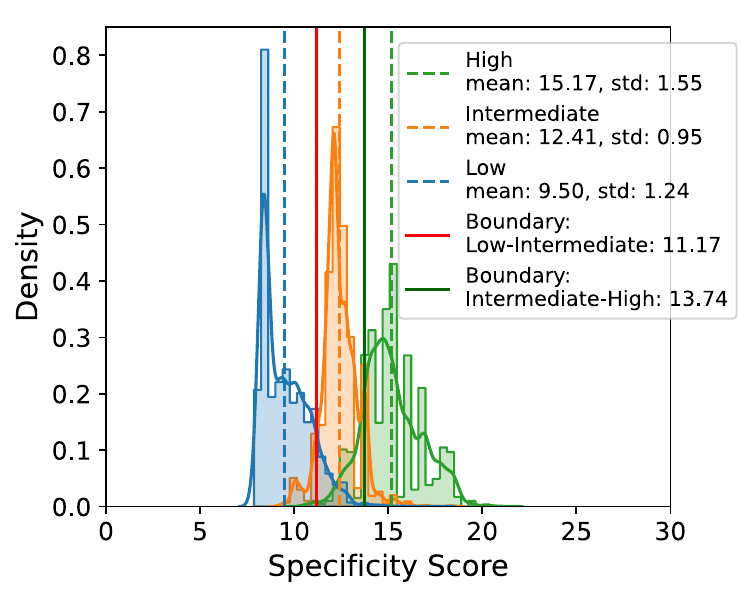} 
        \caption{\textbf{Verbs.}}
        \label{fig:boundv}
    \end{subfigure}
    \caption{\textbf{Specificity Score Distributions for nouns and verbs according to specificity.} The histogram represents the
distribution, including Kernel-Density-Estimations, of the specificity score for nouns and verbs according to their specificity. Based on the intersection points of the Kernel-Density estimations, we derive the specificity level boundaries $B_{\text{Low-Intermediate}} = 17.74$ and $B_{\text{Intermediate-High}} = 20.90$ for nouns, and $B_{\text{Low-Intermediate}} = 11.17$ and $B_{\text{Intermediate-High}} = 13.74$ for verbs.}  
    \label{fig:spec_distr}  
\end{figure}
\begin{multicols}{2}
\section{Experimental Results and Discussion}
The data, processed using the specificity-based synonymization framework, together with the models, are employed to investigate the distributional properties of parts of speech at different specificity levels and the relationship between specificity changes and LLM performance.

\textit{\textbf{Q1}: What is the distribution of specificity scores for each considered part of speech? What are the ranges of specificity scores for low, intermediate and high specificity?}

\textbf{Answer.} We use all the specificity score calculated with eq. \ref{eq:specmes} for nouns and verbs, representing the distribution in Fig. \ref{fig:spec_distr_nv}. In total there were 35848 nouns and 8323 verbs used for this representation. To quantify whether the underlying distributions are normally distributed, we conduct a Kolmogorov-Smirnov test (\cite{doi:https://doi.org/10.1002/9781118445112.stat06558}), which reveals that they are not similar to a normal distribution as seen in Tab. \ref{tab:kstest}. The observable shift, also represented by the strong deviating means of nouns (21.37) and verbs (12.09) specificities, can be explained by minor structural differences in the taxonomy of nouns and verbs. Besides the fact that the overall taxonomy of nouns is larger (82000 unique nouns compared to 11500 unique verbs), the number of direct and indirect hyponyms varies strongly for these two parts of speech, leading to smaller values in the second term of eq. \ref{eq:specmes} for verbs.\\
For the set of adjectives (790 in total), the specificity, as defined by eq. \ref{eq:adjspec}, is predominantly concentrated around the mid-range (mean of 0.40) on a scale from zero to one. This suggests that highly specific adjectives are scarcely represented in our sample. Similar to the distributions observed for nouns and verbs, the specificity of adjectives does not follow a normal distribution, as indicated by the Kolmogorov-Smirnov test, which yielded a p-value of 1.08e-5. It is important to note that the specificity ranges differ between adjectives and the other parts of speech. For adjectives, specificity is calculated using a fractional measure constrained between zero and one, unlike the measures used for nouns and verbs\\
To address the second question, we group the specificity scores by the specificity levels (low, intermediate and high) and combine their individual distribution in Fig. \ref{fig:spec_distr}. Adjectives are excluded from this part of the analysis, since only high specificity synonyms were selected due to the small number of samples remaining after the preprocessing. By categorizing synonyms based on their minimum, average, and maximum values within the pool for a given part of speech, we expect each specificity level's distribution to shift progressively to the right as specificity increases. This expectation is confirmed, as the grouped means consistently rise with increasing specificity levels (the difference in grouped means for nouns and verbs is approximately 3). Notably, the high-specificity group for nouns exhibits a much larger standard deviation (8.26) compared to the low-specificity (2.58) and intermediate-specificity (2.69) groups. This larger variation is likely due to the higher frequency of nouns with specificity scores exceeding 30, which skews the distribution to the right\\
\end{multicols}
\begin{table}[ht!]
    \centering
    \begin{tabular}{|c|c|c|c|c|}
    \hline
    \multirow{2}{*}{Specificity}& \multicolumn{2}{c|}{Nouns} & \multicolumn{2}{c|}{Verbs}\\ 
    \cline{2-5}
     & Lower Bound & Upper Bound & Lower Bound &Upper Bound\\
    \hline
    \hline
    Low Specificity & 10.130 & 17.74 & 7.87 & 11.17 \\
    Intermediate Specificity & 17.741 & 20.90 & 11.171 & 13.74 \\
    High Specificity & 20.901 & 195.62 & 13.741 & 21.16 \\
    \hline
    \end{tabular}
    \caption{\textbf{Specificity Score Intervals.} Interval boundaries for the specificity categories low, intermediate and high for nouns and verbs.}
    \label{tab:ranges}
\end{table}
\begin{multicols}{2}
To accurately define the valid ranges for each specificity category, we employed Kernel Density Estimation (KDE) (\cite{doi:10.1080/24709360.2017.1396742}), a non-parametric method, given that the specificity distributions for nouns and verbs do not follow a normal distribution. We calculated the intersection points between the low-intermediate and intermediate-high levels, resulting in boundaries of 17.74 and 11.74 for low-intermediate and 20.90 and 13.74 for intermediate-high, respectively for nouns and verbs.
The corresponding intervals for each specificity level are presented in Tab. \ref{tab:ranges}.\\

\textit{\textbf{Q2}: How does the specificity of synonyms affect the performance of the models Llama-3.1-70B-Instruct, Granite-13B-Instruct-V2, Flan-T5-XL and Mistral-Large 2 for question answering and reasoning tasks?}

\begin{figure}[ht!]
    \centering
   \includegraphics[width=\textwidth]{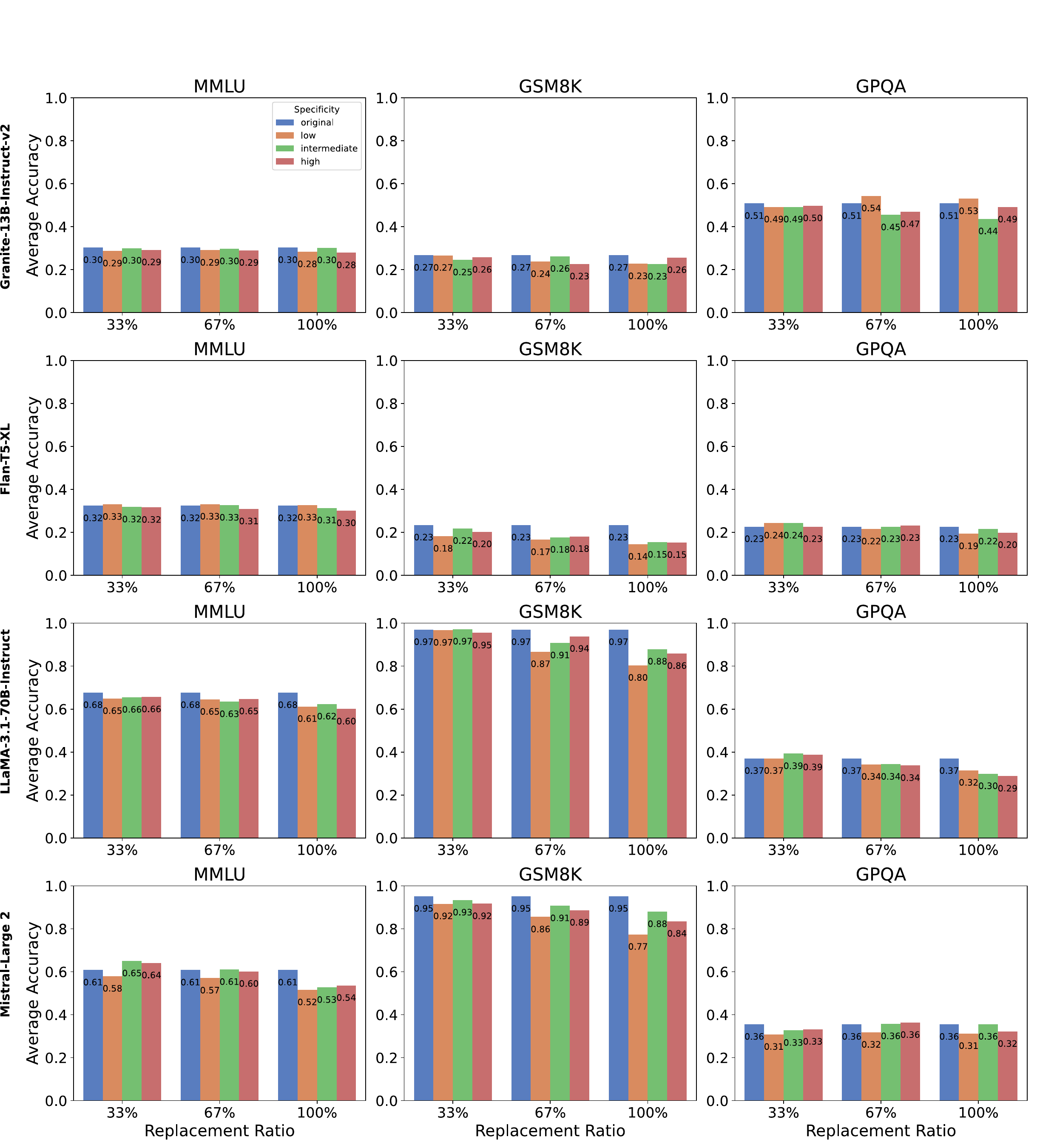}
    \caption{\textbf{Average Accuracy Comparison across multiple LLMs for Nouns.} Average accuracy comparison across all models (Granite-13B-Instruct-v2, Flan-T5-XL, LLaMA-3.1-70B-Instruct, and Mistral-Large 2) for the datasets (MMLU, GSM8K, GPQA) for varying specificity levels (low, intermediate, high) and replacement levels of synonymizable nouns (33\%, 67\%, 100\%).}
    \label{fig:barplot_nouns}
\end{figure}

We separate the analysis of the effects of prompt specificity on the LLM performance in two approaches. The first approach, called \textit{categorical approach}, provides a high-level view by grouping and categorizing nine permutations of specificity levels (low, intermediate, high) and replacement ratios (33\%, 67\%, 100\%) into ordinal categories (33\% Low, ..., 100\% High). It allows for a clearer comparison between these predefined categories with respect to the model performance, and provide insights about potential patterns in how performance shifts across the varying replacement-specificity permutations.\\
The second approach, called \textit{numerical approach}, translates these ordinal replacement-specificity permutations into continuous values, which allows for a more precise quantification of prompt specificity and performance correlation. By treating prompt specificity as a continuous variable and computing correlations, this method captures finer details of how incremental changes in specificity affect performance.\\
\textbf{Categorical Approach.}
For each model, dataset, specificity and replacement level combination we group the outputs and evaluate them with the ground truth. For the evaluation, we opted to use the Jaccard metric (\cite{niwattanakul2013using}) rather than the commonly employed exact-match approach for multiple-choice question-answering tasks. This decision was made to account for responses that included additional characters, such as "A." or "- B," which would incorrectly result in a score of 0 under exact matching, despite the answer being correct. For each combination, we aggregate the Jaccard similarity scores and calculate the average over all samples per category to get the accuracy. Fig. \ref{fig:barplot_nouns} illustrates the comparison of these accuracies for original, low, intermediate and high specificity next to each other for each model, dataset and replacement level. Each row represents the different models and each column shows performance across the datasets: MMLU, GSM8K and GPQA. The y-axis reflects the calculated accuracy, while the x-axis encodes the replacement ratio of the synonymizable nouns. Across all models and datasets, accuracy seems to decrease with an increasing replacement and specificity level, although there are some exceptions from this observable trend. For the GSM8K dataset, the models Granite-13B-Instruct-V2 and Flan-T5-XL show small sensitivity of performance with specificity changes, while for the GPQA dataset, the
performance decreases notably compared to the baseline (original). In particular, for the larger models, Llama-3.1-70B-Instruct and Mistral-Large 2, this performance decrease is strongly visible. Similar trend can be also seen for verbs in the Appendix \ref{sec:appendixa}, Fig. \ref{fig:barplot_verbs}.\\
For adjectives, considering a full replacement with high specificity synonyms, we observe similar performance changes as for the other parts of speech. Notably, the performance of Llama-3.1-70B-instruct and Mistral-Large 2 increases strongly for the GSM8K dataset using high specificity synonyms, which would mark a difference to the other parts of speech. However, the sample size, particularly for GSM8K with 4 instructions, is too small to consider these results as significant. More samples would be required to make a more sophisticated observation.\\
This overall trend indicates that the fractional adjustment of nouns and verbs with higher specificities harmfully affects the LLMs performance, particularly Llama-3.1-70B-Instruct and Mistral-Large 2, in the domains of STEM, law and medicine and therefore seems not to be suitable approach to generally optimize prompts in these domains.
\end{multicols}
\clearpage
\clearpage
\begin{figure}[ht!]
   \includegraphics[width=\textwidth]{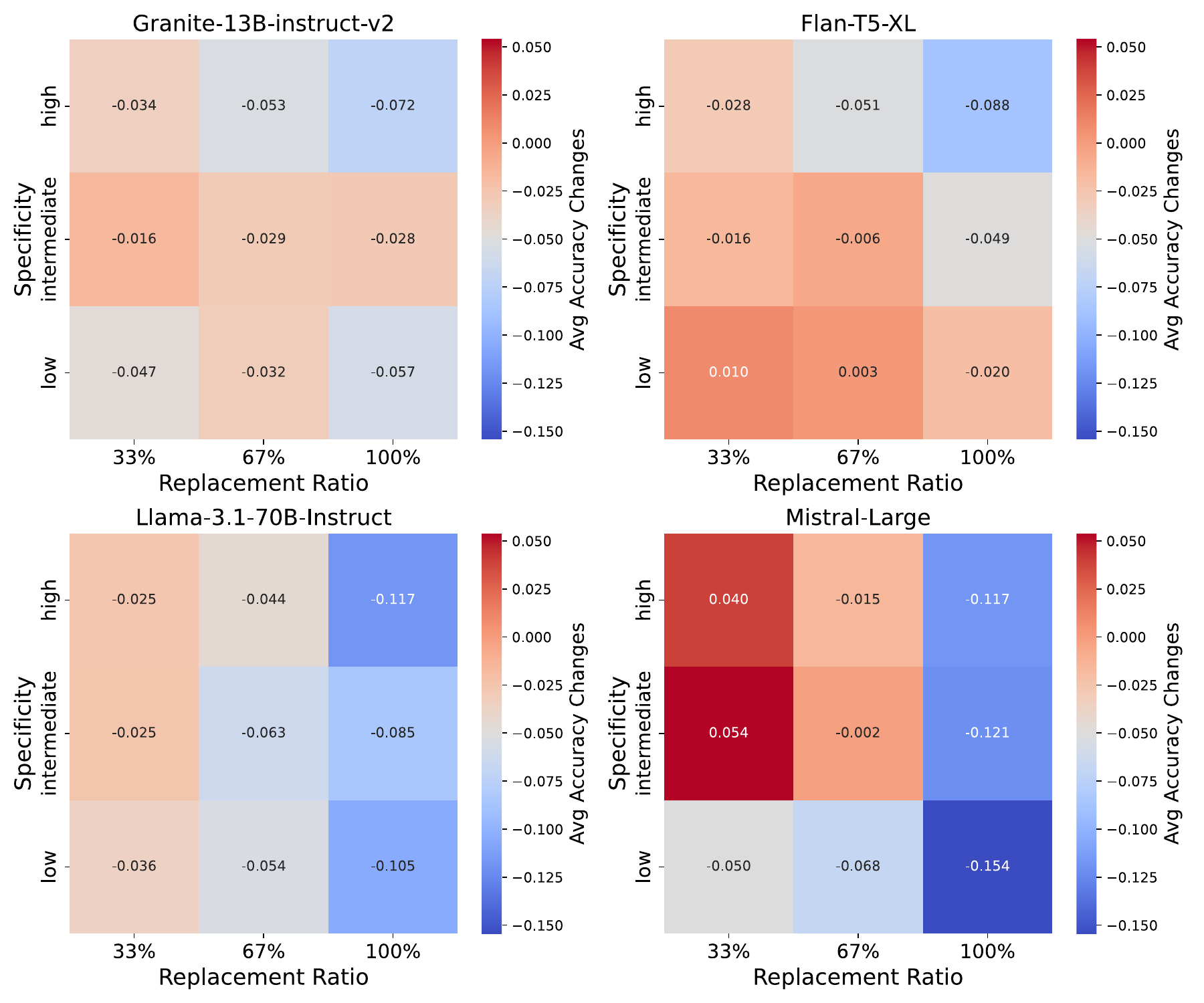}
    \caption{\textbf{Average Accuracy across all Datasets and Models for Nouns.} These heatmaps represent the performance differences in percentages for each model across all data for nouns. Each x-axis represents the replacement ratio (fraction of how many synonyms were used), the y-axis encodes the categorical specificity levels (low, intermediate, high) and the color encodes the average accuracy scores.}
    \label{fig:hm_all_nouns}
\end{figure}
    
\begin{figure}[ht!]
   \includegraphics[width=\textwidth]{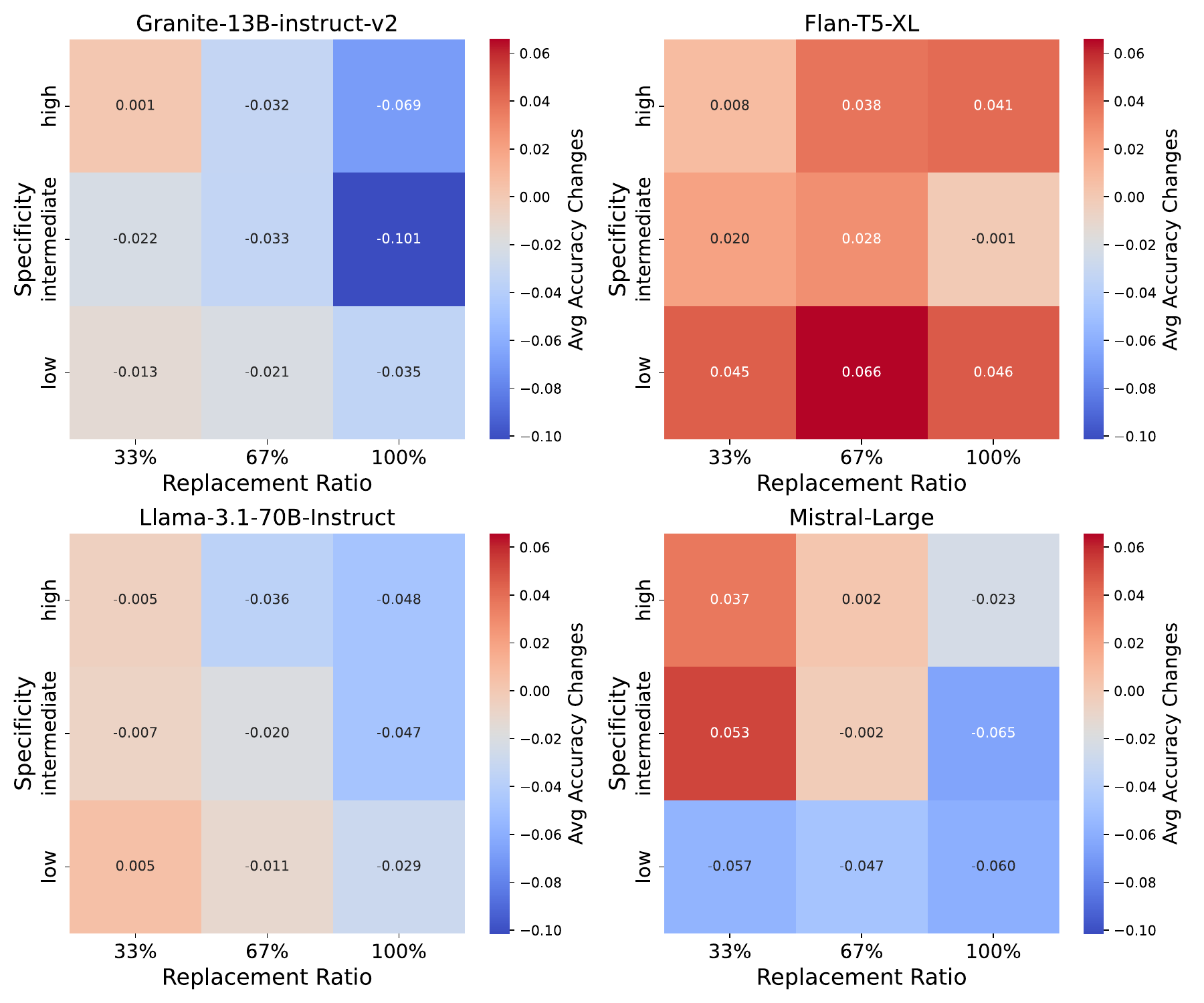}
    \caption{\textbf{Average Accuracy across all Datasets and Models for Verbs.} These heatmaps represent the performance differences in percentages for each model across all data for verbs. Each x-axis represents the replacement ratio (fraction of how many synonyms were used), the y-axis encodes the categorical specificity levels (low, intermediate, high) and the color encodes the average accuracy scores.}
    \label{fig:hm_all_verbs}
    \end{figure}
 \begin{multicols}{2}
To get a better comparison of the performance changes for each combination of specificity and replacement level, we group each class by each model and average the performance across all datasets. We calculate the relative performance change with respect to the baseline performance for each model. The heatmaps in Figs. \ref{fig:hm_all_nouns} and \ref{fig:hm_all_verbs}, respectively for nouns and verbs, illustrate these precentral changes in accuracy when varying specificity levels (low, intermediate, high) and replacement ratios (33\%, 67\%, 100\%) across our four models. Each cell represents the change in performance relative to the baseline (original) accuracy, showing how accuracy shifts when more specific synonyms replace the original words.\\
For nouns, as the replacement ratio and specificity increase, performance generally declines across all models. This indicates that more specific or frequent replacements of synonyms negatively impact the models' abilities to handle zero-shot and chain-of-thought question-answering tasks. For instance, Mistral-Large 2 exhibits the most significant performance drop at 100\% noun replacement and low specificity, showing a decrease of up to -0.154 in accuracy. Llama-3.1-70B-Instruct also shows substantial reductions, especially at higher replacement ratios (-0.117 at high specificity for 100\% replacement). Granite-13B and Flan-T5-XL follow similar patterns, with negative accuracy changes, although Flan-T5-XL shows slightly better resilience to noun replacement.\\
In the case of verbs, the models display more varied responses to synonym specificity and replacement ratios. Flan-T5-XL, for example, demonstrates some positive accuracy changes at lower replacement ratios and low specificity, indicating that it handles verb replacements better, particularly in low and intermediate specificity settings. This suggests that for zero-shot and chain-of-thought question-answering tasks, Flan-T5-XL is more adaptable to verb changes compared to nouns. Mistral-Large also exhibits resilience, showing smaller accuracy reductions for verb replacements, although its performance declines as replacement ratios increase. However, at 100\% verb replacement, all models show a decline in performance, with Granite-13B and Llama-3.1-70B displaying the most significant accuracy drops, particularly at higher specificity levels.

\textbf{Answer.} Overall, the results suggest that noun replacements with more specific synonyms have a stronger negative impact on model performance than verb replacements, with performance generally deteriorating as the replacement ratio increases. The findings highlight that while all models experience degradation in accuracy, Llama-3.1-70B-Instruct and Mistral-Large 2 are more sensitive to high replacement ratios, particularly with specific nouns, whereas Flan-T5-XL exhibits more robustness, especially in handling verb replacements in question-answering tasks. Notably, Llama-3.1-70B-Instruct (0.97) and Mistral-Large 2 (0.95) perform stronger on the reasoning Tasks (GSM8K) compared to Granite-13B-Instruct-V2 and Flan-T5-XL. These performance differences could be attributed to the additional Direct Preference Optimization (Llama-3.1)(\cite{NEURIPS2023_a85b405e}) and larger parameter size (Mistral-Large 2 with 123B parameters) (\cite{kaplan2020scalinglawsneurallanguage}). However, the categorical approach only offers a broach perspective on the potential relationship between increasing specificity and LLM performance. By only focusing on the predefined categories, specificity level and replacement ratio, this approach does not take into account the overall prompt specificity since instructions are assigned to one category based on the number of individual synonyms with varying specificity. There might be cases where a modified instruction will be assigned to the 33\% High 
\end{multicols}
\begin{table}[H]
\centering
\adjustbox{max width=\textwidth}{
\begin{tabular}{|c|c|c|c|c|c|c|}
\hline
\multirow{3}{*}{Model} & \multicolumn{6}{|c|}{Nouns} \\ \cline{2-7}
 & \multicolumn{2}{|c|}{MMLU\textit{(zero-shot)}} & \multicolumn{2}{c|}{GPQA\textit{(zero-shot)}} & \multicolumn{2}{c|}{GSM8K\textit{(zero-shot, CoT)}} \\ \cline{2-7}
 & Corr & p-value & Corr & p-value & Corr & p-value\\ \hline
 Granite-13b-instruct-v2 & -0.48 & 0.159 & \textbf{\underline{-0.68}} & \textbf{0.030} & -0.20 & 0.580\\ \hline
Flan-T5-XL&0.36 &  0.309 & 0.04 & 0.906 & -0.04 & 0.907 \\ \hline
Llama-3.1-70b-instruct& -0.29 & 0.422 & -0.32 & 0.365 & -0.001 & 0.987 \\ \hline
Mistral-Large 2& -0.05 & 0.881 &0.62 & 0.054 & -0.01 & 0.987 \\ \hline
 \hline
 \multirow{3}{*}{Model} & \multicolumn{6}{c|}{Verbs} \\ \cline{2-7}
  & \multicolumn{2}{|c|}{MMLU\textit{(zero-shot)}} & \multicolumn{2}{c|}{GPQA\textit{(zero-shot)}} & \multicolumn{2}{c|}{GSM8K\textit{(zero-shot, CoT)}} \\ \cline{2-7}
   & Corr & p-value & Corr & p-value & Corr & p-value\\ \hline
 Granite-13b-instruct-v2 & -0.37 & 0.292 & 0.46 & 0.177 & -0.18 & 0.627 \\ \hline
Flan-T5-XL& -0.08 & 0.836 & 0.50 & 0.139 & \textbf{-0.78} & \textbf{0.008} \\ \hline
Llama-3.1-70b-instruct& \textbf{-0.79} & \textbf{0.006} &0.42 & 0.229 & \textbf{\underline{-0.89}} & \textbf{5.4e-4} \\ \hline
Mistral-Large 2& 0.38 & 0.276 & -0.02 & 0.960 & \textbf{-0.87} & \textbf{0.001} \\ \hline

\end{tabular}
}
\caption{\textbf{Correlation Results.} Results (Spearman Correlation) for each model, grouped on each part of speech and each dataset. Values in \textbf{bold} display the significant correlations, where the p-value is under 0.05, while \underline{underlined} values represent the greatest correlation out of all significant ones.}
\label{tab:corr}
\end{table}

\begin{multicols}{2}
category, but would exhibit a smaller prompt specificity than the 100\% intermediate variation. To address this limitation of the categorical approach, we shift from a word-level focus to a prompt specificity perspective in the \textit{numerical approach}.

\textbf{Numerical Approach.}
For each model, we systematically aggregate all outputs alongside the corresponding ground truth labels provided by the original datasets. We then partition these into sub-datasets based on part of speech (nouns and verbs) and dataset type (MMLU, GSM8K, GPQA), resulting in six distinct sub-datasets per model. To evaluate the performance of the LLM, we employ the Jaccard metric as utilized in the \textit{categorical approach}. Final performance scores for each dataset and model are obtained by calculating accuracy across all samples, stratified by nine permutations of specificity levels (low, intermediate, high) and replacement levels (33\%, 67\%, and 100\%).\\
Additionally, for each modified instruction, we compute the prompt specificity corresponding to the analyzed part of speech (refer to Section \ref{sec:promptspec}), and average these scores across each combination of specificity and replacement levels within each dataset. This comprehensive approach allows us to capture the holistic effect of synonymization with varying specificities on the entire prompt, rather than focusing solely on the subset of synonyms. For instance, in a prompt containing ten nouns, where only three nouns are subject to synonymization, calculating the average specificity based only on the three modified nouns may yield skewed results compared to averaging across all ten nouns. Such an approach could artificially inflate the perceived impact of synonymization, even though the actual changes might be minimal. Furthermore, while discrepancies in total word count between two samples with an equivalent number of nouns could introduce bias, we deem this effect negligible, as the number of nouns is likely to positively correlate with the number of sentences, thus diminishing the bias in our approach.\\
Finally, we compute the Spearman correlation between the average prompt specificity and the performance of the LLM across datasets and specificity levels to evaluate the relationship between these variables. We use Spearman instead of Pearson, since there are only ten values for each parameter, and we cannot assume that these are normally distributed.

\textbf{Answer.} Table \ref{tab:corr} presents a comparison of correlations and p-values for four large LLMs across three distinct tasks, MMLU (zero-shot), GPQA (zero-shot), and GSM8K (zero-shot, Chain-of-Thought), separated by nouns and verbs. In general, increasing prompt specificity tends to negatively impact LLM performance, though most of these changes are not statistically significant (p-values \(> 0.05\)). For nouns, the only significant result is a strong negative correlation of \(-0.68\) (p-value = 0.03) observed for Granite-13b-instruct-v2 on the GPQA (zero-shot) task. For verbs, significant negative correlations are observed for Llama-3.1-70b-instruct on MMLU \((-0.79\), p = 0.006) and Flan-T5-XL on GSM8K \((-0.78\), p = 0.008), along with very strong negative correlations for GSM8K with Llama-3.1-70B-Instruct \((-0.89\), p = 5.4e-4) and Mistral-Large 2 \((-0.87\), p = 0.001). Notably, most of the significant correlations pertain to verbs. However, with only five out of 24 evaluations showing significant results, this suggests that synonymization generally does not have a substantial impact on LLM performance for our tasks.\\
\end{multicols}
\justifying
\begin{figure}[H]
    \centering
     \begin{subfigure}[b]{0.45\textwidth}
       \includegraphics[width=\textwidth]{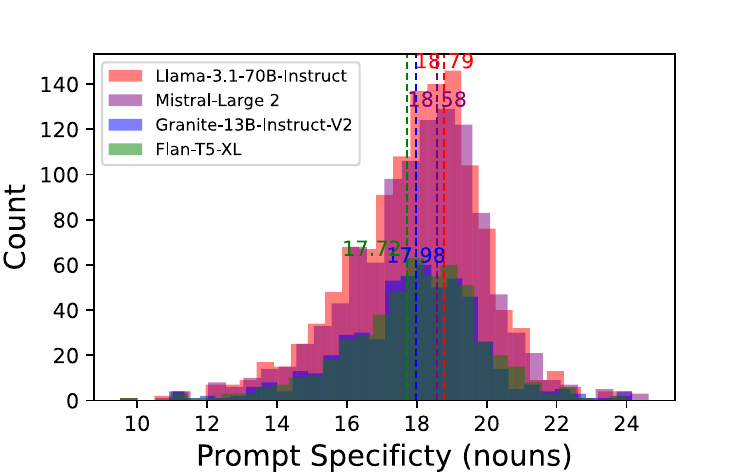}
        \caption{\textbf{Distribution of correct answers for nouns.}}
        \label{fig:original_distr_n}
    \end{subfigure}
    \hspace{0.05\textwidth}
    \begin{subfigure}[b]{0.45\textwidth}
   \includegraphics[width=\textwidth]{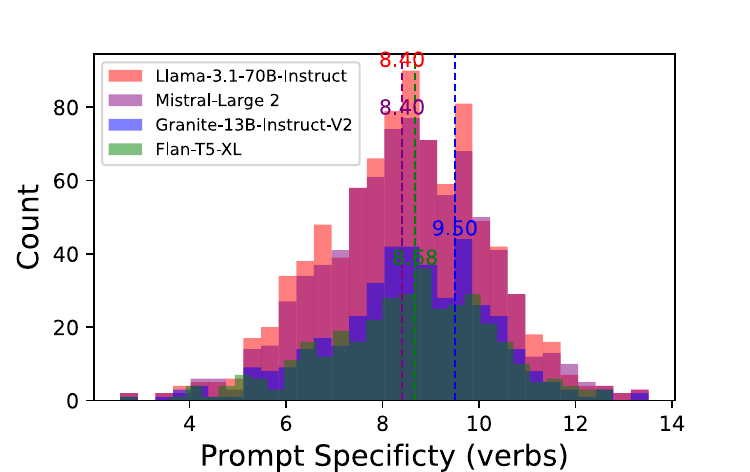}
    \caption{\textbf{Distribution of correct answers for verbs.} }
    \label{fig:original_distr_v}
    \end{subfigure}
    \caption{\textbf{Distribution of correct Answers for each Model.} These histograms represent the distributions of correct answers for the original samples for each model and part of speech. The x-axis depicts the respective prompt specificity of the unmodified instructions, while the y-axis counts the correct answers for the corresponding prompt specificity.}
    \label{fig:original_distr}
\end{figure}
\begin{multicols}{2}
Contradictory to our intuition, that increasing the prompt specificity would lead to better LLM performance, these results imply a rather unexpected opposite outcome. However, this evaluation reveals several insights into how increasing prompt specificity impacts LLM performance. Although the overall trend indicates a negative effect on performance, the fact that most changes are not statistically significant suggests that LLMs are generally resilient to prompt variations in terms of specification through synonymization.\\
Interestingly, the higher frequency of significant negative correlations observed for verbs in the reasoning task (GSM8K) compared to nouns indicate that prompt changes related to actions and relationships in a sentence (verbs) may have a more disruptive effect on model performance in reasoning than changes in descriptive or referential elements (nouns). It is possible that models require different levels of abstraction or generality depending on the dataset's task structure, with reasoning tasks being negatively impacted from more specific terms, as these might complicate the LLMs ability to follow logical progression. In particular, Llama-3.1-70b-instruct and Mistral-Large 2 exhibited very strong negative correlations, suggesting that these models are especially sensitive to prompt variations in tasks requiring problem-solving and reasoning. Their heightened sensitivity to changes in verb specificity could be attributed to differences in their pre-training data distributions or the manner in which they handle syntactic dependencies. Llama-3.1-70B-Instruct and Mistral Large 2 may have been exposed to fewer domain-specific verbs, leading to increased difficulty when processing specific verb synonyms. Further research into the architectural differences between the considered models might reveal whether these sensitivities are due to inherent biases in the attention mechanisms or pre-training corpora.\\
While this model-depended sensitivity might reflect model-specific weaknesses, the small number of significant results overall (5 out of 24 evaluations) suggests that synonymization does not pose a substantial challenge for most LLMs in the tested tasks.\\

\textit{\textbf{Q3.} Is there a specific level of prompt specificity for nouns and verbs that results in the best LLM performance in question answering and reasoning tasks across different models?}

To address this question, we analyzed the outputs of the original, unmodified samples for each model and dataset, applying a stricter evaluation criterion compared to previous analyses by using exact match to identify correct answers. This approach was adopted to eliminate all noise and ensure higher precision in the evaluation. Additionally, we computed the prompt specificity for each part of speech (nouns and verbs) in the original prompts. The aggregation of correct answers across all datasets for each model is visualized in Fig. \ref{fig:original_distr}. In each histogram, the x-axis represents the prompt specificity for the respective part of speech, while the y-axis indicates the count of correct answers. A dashed line marks the prompt specificity associated with the highest number of correct answers for each model.\\
For nouns, the distributions converge around similar prompt specificity values, ranging from 17.72 to 18.79. The more recent models, Llama-3.1 and Mistral-Large 2, exhibit higher median specificities, whereas Flan-T5-XL and Granite-13B achieve optimal performance with less specific prompts. This indicates a positive correlation between higher prompt specificity for nouns and 
\end{multicols}
\begin{table}[ht]
\centering
\adjustbox{max width=\textwidth}{
\begin{tabular}{|c|c|c|c|c|c|c|}
\hline
\multirow{2}{*}{Model} & \multicolumn{3}{|c|}{Nouns} & \multicolumn{3}{|c|}{Verbs} \\ \cline{2-7}
 & Lower & Upper & Median & Lower & Upper & Median\\ \hline
 Granite-13b-instruct-v2 & 17.24 & 22.00 & 18.95 & 9.20 & 14.74  & 10.55\\ \hline
Flan-T5-XL& 17.03& 22.04  & 18.94 & 8.29 & 14.80 & 10.47 \\ \hline
Llama-3.1-70b-instruct& 17.58 & 22.57 & 19.41 & 8.08 & 14.14 & 10.57 \\ \hline
Mistral-Large 2& 17.58 & 22.57 & 19.70 & 9.03 & 14.49 & 10.57 \\ \hline
\end{tabular}
}
\caption{\textbf{Optimal Specificities.} Ranges of prompt specificities for nouns and verbs, in which the LLM perform best across all datasets.}
\label{tab:optimalranges}
\end{table}
\begin{multicols}{2}
the number of correct answers; however, beyond a certain threshold, increasing specificity may not enhance performance across all models and could, in fact, negatively impact accuracy.\\
A similar pattern is observed for verbs, with the highest concentration of correct answers corresponding to prompt specificities between 8.4 and 9.5. Granite-13B performs best with a higher verb specificity (9.5), while other models achieve their optimal performance at slightly lower prompt specificities.\\
From these observations, we can infer that for the original, unmodified samples, a moderate level of specificity improves model performance. However, excessive or insufficient specificity may degrade performance across models. We apply this approach to the modified instructions, examining each replacement level, specificity level, model, and dataset combination. The resulting prompt specificities for each parameter combination are displayed in Appendix \ref{sec:appendixb}, Tab. \ref{tab:maxes}.

\textbf{Answer.} Based on this assessment, we derive the ranges of optimal prompt specificity for each model and part of speech, displayed in Tab. \ref{tab:optimalranges}, where the LLM achieves the highest performance across all datasets.
The results indicate that there is a range of prompt specificity for both nouns and verbs, beyond which further decreases or increases in specificity can negatively impact performance. For nouns, the optimal specificity falls between 17.72 and 19.70, with newer models like Llama-3.1 and Mistral-Large 2 performing better with slightly higher specificity than smaller models like Granite-13B and Flan-T5-XL. Similarly, for verbs, optimal performance occurs between 8.08 and 10.57, with Granite-13B benefitting from higher specificity compared to other models. These findings suggest that while some level of specificity improves accuracy, exceeding the optimal range can diminish performance, highlighting the importance of fine-tuning prompt specificity based on the model and task. Particularly, in domain-specific NLP applications, one should consider that increasing prompt specificity, particularly for verbs, may not always yield better performance. Instead, balancing specificity with generality could help models retain broader reasoning capabilities while maintaining accuracy.
\section{Conclusion}
In this study, we examined the effects of synonymizing nouns, verbs, and adjectives with varying levels of specificity across four different models. By applying this synonymization to samples from three domain-specific datasets for question-answering and reasoning tasks, we identified a range of prompt specificity for nouns and verbs, consistent across all models, that leads to the best LLM performance for these tasks.\\
Contrary to our initial hypothesis that increasing prompt specificity would improve LLM outputs, our findings show no significant performance changes when increasing noun specificity in prompts. However, using more specific verb synonyms resulted in a negative impact on performance in reasoning tasks. This suggests that prompt design might benefit more from focusing on clarity, contextual appropriateness and other linguistic factors rather than purely on specificity.\\
We also introduced an approach to quantify adjective specificity that shows an overall intermediate-strong correlation with the ranking of GPT-4o. However, our experiments showed that adjectives are underrepresented in our question answering datasets and are not suitable for further analysis to quantify their impact on LLM performance in our scope. Despite that, it is plausible that adjectives play a more critical role in tasks involving subjective analysis, such as sentiment analysis, where they carry significant emotional or descriptive meaning. Therefore, future research could explore how adjective specificity affect the LLM performance in other NLP tasks, such as sentiment analysis or narrative text generation, where they have a more substantial role compared to the neutral toned question answering tasks.
\clearpage
\section{Limitations}
Despite the promising results obtained from our exploration of specificity affects LLM performance, several limitations remain that warrant further attention.\\
First, our approach relies on WSD to identify and substitute synonyms. Although, our algorithm guarantees the correct sense of a word most of the time, as shown in Tab. \ref{tab:wsd}, it misclassifies the sense in some cases. This limitation can result in the selection of inappropriate synonyms with respect to the given context, which may cause confusion for the model and affect performance by introducing noise into the input and reducing the precision of the generated outputs. In the future, one could improve the WSD process by fine-tuning Llama-3.1-70B-Instruct specifically on the WSD task similar to \cite{wahle2021incorporating}, since the base version already provided promising results (79\% accuracy). A less computational resource-intensive method to address this issue would be to incorporate human feedback after the WSD process, which would allow a manual correction of the false senses.\\
Second, our analysis primarily focused on the domains of STEM, medicine, and law. While these areas are widely applicable, they do not fully encompass the breadth of all potential fields and contexts in which domain-oriented prompt engineering may be applied. Consequently, the findings in this work may not generalize to other domains, limiting the overall applicability of the results. Future research could extend the scope to include a wider variety of domains such as psychology, finance or engineering, to enhance the robustness of the conclusions.
Finally, the introduction of an adjective specificity equation represents an initiative toward quantifying and ranking adjectives. Although, this equation shows promising results according to our evaluation in the STEM, law and medicine domains, its effectiveness needs to be rigorously tested on large scale across a larger scope of domains to determine its generalizability.
\section{Acknowledgments}
I would like to extend my deepest gratitude to Jan Philip Wahle and Dr. Terry Ruas for their invaluable feedback, insightful perspectives, and continuous support throughout the course of my thesis. I am also profoundly thankful to Prof. Dr. Bela Gipp for the opportunity to conduct this research under his supervision, enabling me to contribute to the NLP and Generative AI research community.\\
Additionally, I would like to acknowledge IBM for providing access to their Generative AI infrastructure, which was integral to our experimental work.\\
Finally, I am sincerely grateful to my friends and family, particularly Jana Fedjaev, for their steadfast support during this intense phase of my life.\\

In the conduct of this research project, we used specific artificial intelligence tools and algorithms GPT-4o to assist with rephrasing sentences. While these tools have augmented our capabilities and contributed to our findings, it's pertinent to note that they have inherent limitations. We have made every effort to use AI in a transparent and responsible manner. Any conclusions drawn are a result of combined human and machine insights. This is an automatic report generated with © AI Usage Cards https://ai-cards.org
\printbibliography
\end{multicols}

\clearpage
\appendix
\section{Figures}
\label{sec:appendixa}
\begin{figure}[ht!]
\definecolor{lightred}{HTML}{FF9694}
\definecolor{lightblue}{HTML}{6eb2fa}
    \centering
   \includegraphics[width=\textwidth]{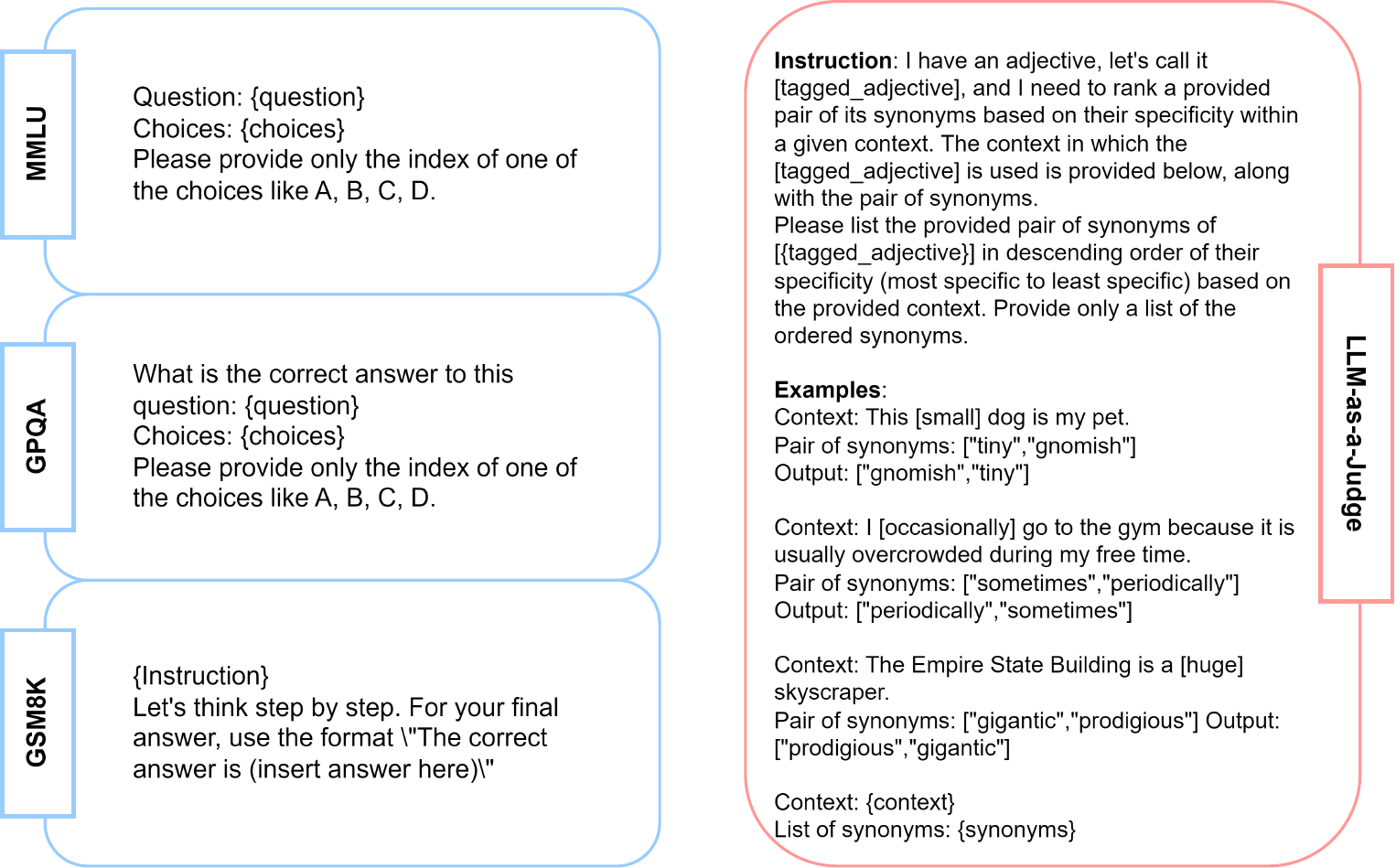}
    \caption{\textbf{Prompt Templates.} Overview of all the prompt templates used for performing the question-answering and reasoning tasks (\textcolor{lightblue}{MMLU, GPQA and GSM8K}), and the \textcolor{lightred}{LLM-as-a-Judge} experiment for the adjective specificity measure evaluation.}
    \label{fig:prompts}
\end{figure}
\begin{figure}[ht!]
    \centering
   \includegraphics[width=\textwidth]{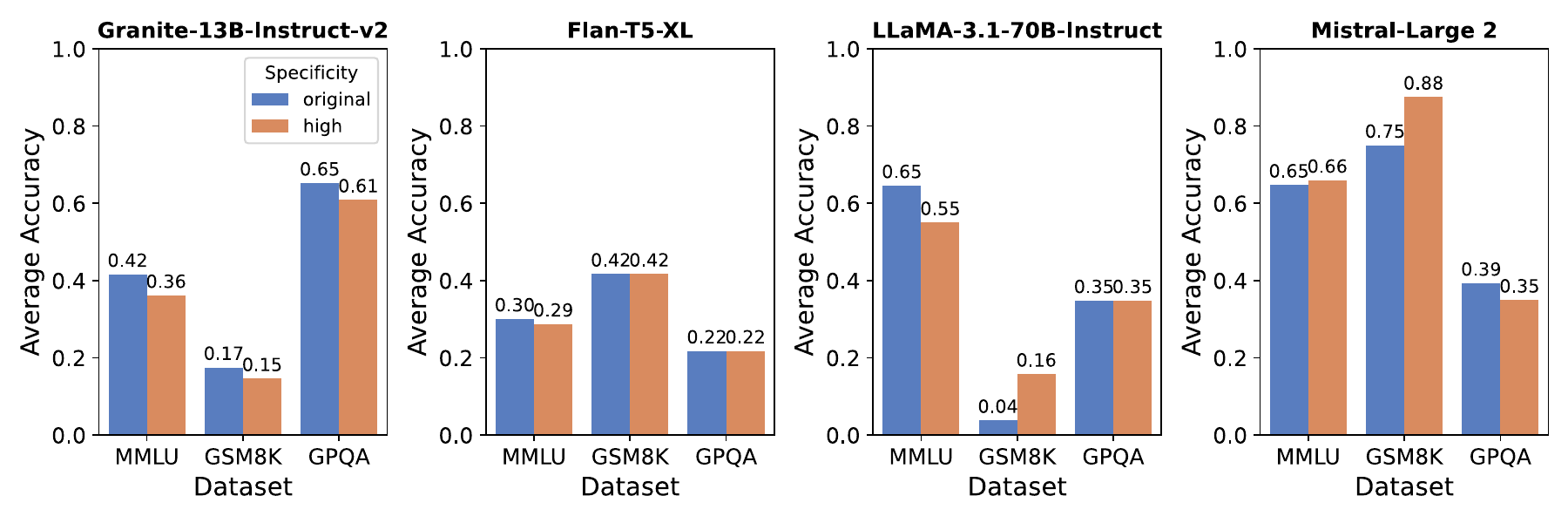}
    \caption{\textbf{Average Accuracy Comparison across multiple LLMs for Adjectives.} Average accuracy comparison across all models (Granite-13B-Instruct-v2, Flan-T5-XL, LLaMA-3.1-70B-Instruct, and Mistral-Large 2) for the datasets (MMLU, GSM8K, GPQA) for original and 100\% high specificity instructions. Due to the small sample size of 172 instructions in total, these results cannot be considered as significant.}
    \label{fig:barplot_a}
\end{figure}
\begin{figure}[ht!]
    \centering
   \includegraphics[width=\textwidth]{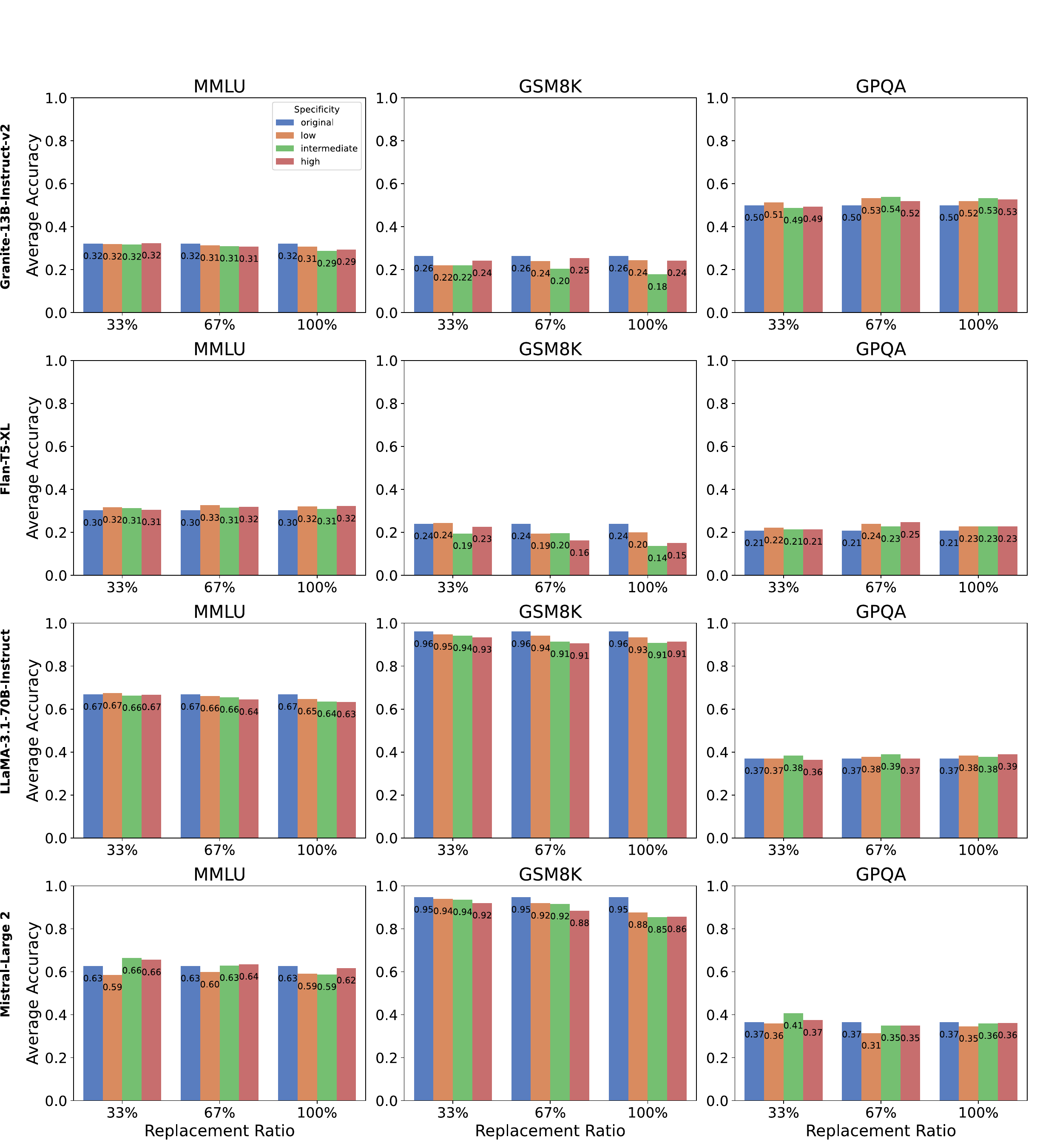}
    \captionof{figure}{\textbf{Average Accuracy Comparison across multiple LLMs for Verbs.} Average accuracy comparison across all models (Granite-13B-Instruct-v2, Flan-T5-XL, LLaMA-3.1-70B-Instruct, and Mistral-Large 2) for the datasets (MMLU, GSM8K, GPQA) for varying specificity levels (low, intermediate, high) and replacement levels of synonymizable verbs (33\%, 67\%, 100\%).}
    \label{fig:barplot_verbs}
\end{figure}
\clearpage
\section{Tables}
\label{sec:appendixb}
\begin{table}[ht!]
\centering
\adjustbox{max width=\textwidth}{
\begin{tabular}{|c|c|c|c|c|c|c|c|c|}
\hline
\multirow{2}{*}{Dataset} & \multirow{2}{*}{Model} & \multirow{2}{*}{Replacement Ratio [\%]} & \multicolumn{3}{|c|}{Nouns} & \multicolumn{3}{|c|}{Verbs} \\ \cline{4-9}
 & & & Low & Intermediate & High & Low & Intermediate & High\\ \hline
 \hline
  \multirow{12}{*}{\rotatebox{90}{All}}&\multirow{3}{*}{Granite-13b-instruct-v2} & 33 & 17.23 & 18.49 & 18.95 & 9.20 & 9.96 & 10.55 \\ \cline{3-9}
 && 67 & 17.58 & 19.24 & 20.52 & 9.24 & 10.87 & 12.63\\ \cline{3-9}
& & 100 & 17.71 & 19.67 & 22.00 & 9.23 & 12.19 & 14.74\\ \cline{2-9}
&\multirow{3}{*}{Flan-T5-XL} & 33 & 17.03 & 18.56 & 18.94 & 8.29 & 9.98 & 10.47 \\ \cline{3-9}
 && 67 & 17.54 & 19.16 & 20.90 & 9.31 & 10.89 & 12.29 \\ \cline{3-9}
 && 100 & 17.48 & 20.38 & 22.04 & 9.42 & 11.92 & 14.80 \\ \cline{2-9}
 &\multirow{3}{*}{Llama-3.1-70B-Instruct} & 33 & 18.61 & 18.76 & 19.41 & 8.08 & 9.86  & 10.57\\ \cline{3-9}
 && 67 & 17.58 & 19.70 & 20.90 & 9.29 & 10.59 & 12.82\\ \cline{3-9}
 && 100 & 17.83 & 19.66 & 22.57 & 9.03 & 12.14 & 14.49\\ \cline{2-9}
& \multirow{3}{*}{Mistral-Large 2} & 33 & 17.97 & 18.86 & 19.70 & 9.20 & 9.86 & 10.57 \\ \cline{3-9}
& & 67 & 17.58 & 19.70 & 20.58 & 9.29 & 10.59 & 12.82\\ \cline{3-9}
& & 100 & 17.83 & 20.35 & 22.57 & 9.03 & 12.14 & 14.49\\ \hline
\hline
 \multirow{12}{*}{\rotatebox{90}{MMLU}}&\multirow{3}{*}{Granite-13b-instruct-v2} & 33 &  17.24& 18.49 & 18.95 & 9.20 & 9.96 & 10.55\\ \cline{3-9}
 && 67 & 17.58 & 19.24 & 20.52 & 9.24 & 10.87 & 12.63\\ \cline{3-9}
& & 100 & 17.71 & 19.67 & 22.00 & 9.23 & 12.19 & 14.74\\ \cline{2-9}
&\multirow{3}{*}{Flan-T5-XL} & 33 & 17.03 & 18.60 & 18.94 & 8.29 & 9.98 & 10.47\\ \cline{3-9}
 && 67 & 17.54 & 19.16 & 20.90 & 9.31 & 10.89 & 12.29\\ \cline{3-9}
 && 100 & 17.48 & 20.38 & 22.04 & 9.42 & 11.92 & 14.80\\ \cline{2-9}
 &\multirow{3}{*}{Llama-3.1-70B-Instruct} & 33 & 18.61 & 18.76 & 19.41 & 8.08 & 9.86 & 10.57\\ \cline{3-9}
 && 67 & 17.58 & 19.70 & 20.90 & 9.29 & 10.59 & 12.61\\ \cline{3-9}
 && 100 & 17.83 & 19.65 & 22.57 & 9.03 & 12.14 & 14.14\\ \cline{2-9}
& \multirow{3}{*}{Mistral-Large 2} & 33 & 17.97 & 18.86 & 19.70 & 9.20 & 9.86 & 10.57 \\ \cline{3-9}
& & 67 & 17.58 & 19.70 & 20.58 & 9.29 & 10.59 & 12.82\\ \cline{3-9}
& & 100 & 17.83 & 20.35 & 22.57 & 9.03 & 12.14& 14.49\\ \hline
\hline
 \multirow{12}{*}{\rotatebox{90}{GSM8K}}&\multirow{3}{*}{Granite-13b-instruct-v2} & 33 & 19.29 & 19.59 & 20.16 & 8.40 & 7.40 & 9.14 \\ \cline{3-9}
 && 67 & 16.54 & 19.05 & 19.21 & 8.63 & 9.51 & 10.86\\ \cline{3-9}
& & 100 & 18.25 & 19.12 & 23.12 & 9.30 & 10.24 & 12.08\\ \cline{2-9}
&\multirow{3}{*}{Flan-T5-XL} & 33 & 20.86 & 20.29 & 19.16 & 8.20 & 9.80 & 9.51\\ \cline{3-9}
 && 67 & 16.25 & 20.25 & 20.28 & 8.85 & 6.64 & 7.99\\ \cline{3-9}
 && 100 & 16.18 & 17.75 & 23.93 & 8.94 & 10.29 & 14.48\\ \cline{2-9}
 &\multirow{3}{*}{Llama-3.1-70B-Instruct} & 33 & 19.26 & 19.53 & 20.09 & 8.81 & 9.79 & 10.56\\ \cline{3-9}
 && 67 & 18.35 & 19.88 & 21.75 & 8.57 & 9.90 & 11.09\\ \cline{3-9}
 && 100 & 18.51 & 19.66 & 22.03 & 9.22 & 11.36 & 14.60\\ \cline{2-9}
& \multirow{3}{*}{Mistral-Large 2} & 33 & 19.26 & 19.53 & 20.09 & 8.81 & 9.79 & 10.36\\ \cline{3-9}
& & 67 & 18.83 & 19.88 & 21.75 & 8.81 & 9.90 & 11.75 \\ \cline{3-9}
& & 100 & 18.51 & 21.11 & 23.63 & 8.59 & 11.36 & 14.60\\ \hline
\hline
 \multirow{12}{*}{\rotatebox{90}{GPQA}}&\multirow{3}{*}{Granite-13b-instruct-v2} & 33 & 16.96 & 18.58 & 18.38 & 7.97 & 9.77 & 11.13\\ \cline{3-9}
 && 67 & 16.55 & 19.38 & 19.06 & 9.22 & 12.08 & 12.22\\ \cline{3-9}
& & 100 & 16.57 & 19.59 & 20.54 & 8.59 & 11.99 & 14.47\\ \cline{2-9}
&\multirow{3}{*}{Flan-T5-XL} & 33 & 17.05 & 17.29 & 18.45 & 7.86 & 9.18 & 11.40\\ \cline{3-9}
 && 67 & 17.04 & 20.27 & 18.88 & 8.39 & 10.00 & 11.41\\ \cline{3-9}
 && 100 & 16.61 & 19.68 & 22.05 & 9.05 & 12.46 & 14.66\\ \cline{2-9}
 &\multirow{3}{*}{Llama-3.1-70B-Instruct} & 33 & 17.01 & 17.47 & 18.38 & 8.24 & 8.31 & 9.24\\ \cline{3-9}
 && 67 & 17.79 & 17.76 & 19.06 & 9.25 & 10.76 & 11.26\\ \cline{3-9}
 && 100 & 17.71 & 19.58 & 21.02 & 9.37 & 12.33 & 11.85\\ \cline{2-9}
& \multirow{3}{*}{Mistral-Large 2} & 33 & 19.07 & 17.44 & 18.38 & 8.24 & 8.31  & 9.32 \\ \cline{3-9}
& & 67 & 16.92 & 17.98 & 20.81 & 9.25 & 10.13 & 11.12\\ \cline{3-9}
& & 100 & 17.87 & 19.55 & 21.01 & 9.37 & 12.33 & 11.85\\ \hline
\end{tabular}
}
\caption{\textbf{Optimal Specificities for each Dataset.} Optimal prompt specificities for nouns and verbs for each combination of replacement ratios (33\%, 67\%, 100\%) and specificity levels (low, intermediate, high) for all data combined, MMLU, GSM8K and GPQA.}
\label{tab:maxes}
\end{table}

\begin{table}[ht!]
    \centering
    \begin{tabular}{|c|c|}
    \hline
        Parameter & Value \\ \hline \hline
        temperature & 0 \\ \hline
        max\_tokens & 500 \\ \hline
        random\_seed & 31415 \\ \hline
    \end{tabular}
    \caption{\textbf{Model Parameters.} Overview of parameters used for prompting in all the experiments in this study.}
    \label{tab:params}
\end{table}
\clearpage
\section{Data overview}
\label{sec:appendixdata}
\begin{table}[ht!]
    \centering
    \begin{tabular}{|c|c|c|c|c|}
    \hline
         \multicolumn{2}{|c|}{Dataset}& Domains & Difficulty & Samples  \\ \hline\hline
         \multirow{12}{*}{\rotatebox{90}{MMLU}} &task686&biology&college&162\\\cline{3-5}
         &task687&chemistry&college&110\\\cline{3-5}
         &task688&computer science&college&113\\\cline{3-5}
         &task689&mathematics&college&113\\\cline{3-5}
         &task691&physics&college&96\\\cline{3-5}
         &task699&biology&high school&184\\\cline{3-5}
         &task700&chemistry&high school&174\\\cline{3-5}
         &task701&computer science&high school&111\\\cline{3-5}
         &task708&physics&high school&169\\\cline{3-5}
         &task710&statistics&high school&175\\\cline{3-5}
         &task729&law&professional&302\\\cline{3-5}
         &task730&medicine&professional&183\\\hline
         \multicolumn{2}{|c|}{GSM8K} & mathematics & grade school & 7473\\\hline
         \multicolumn{2}{|c|}{GPQA} & biology, chemistry, physics & PhD graduate& 448\\\hline
    \end{tabular}
    \caption{\textbf{Overview of Datasets.} The overview of all datasets used in our experiments with the corresponding domains, difficulty levels and initial sample sizes.}
    \label{tab:dataall}
\end{table}
\begin{figure}[ht!]
    \centering
    \includegraphics[width=\textwidth]{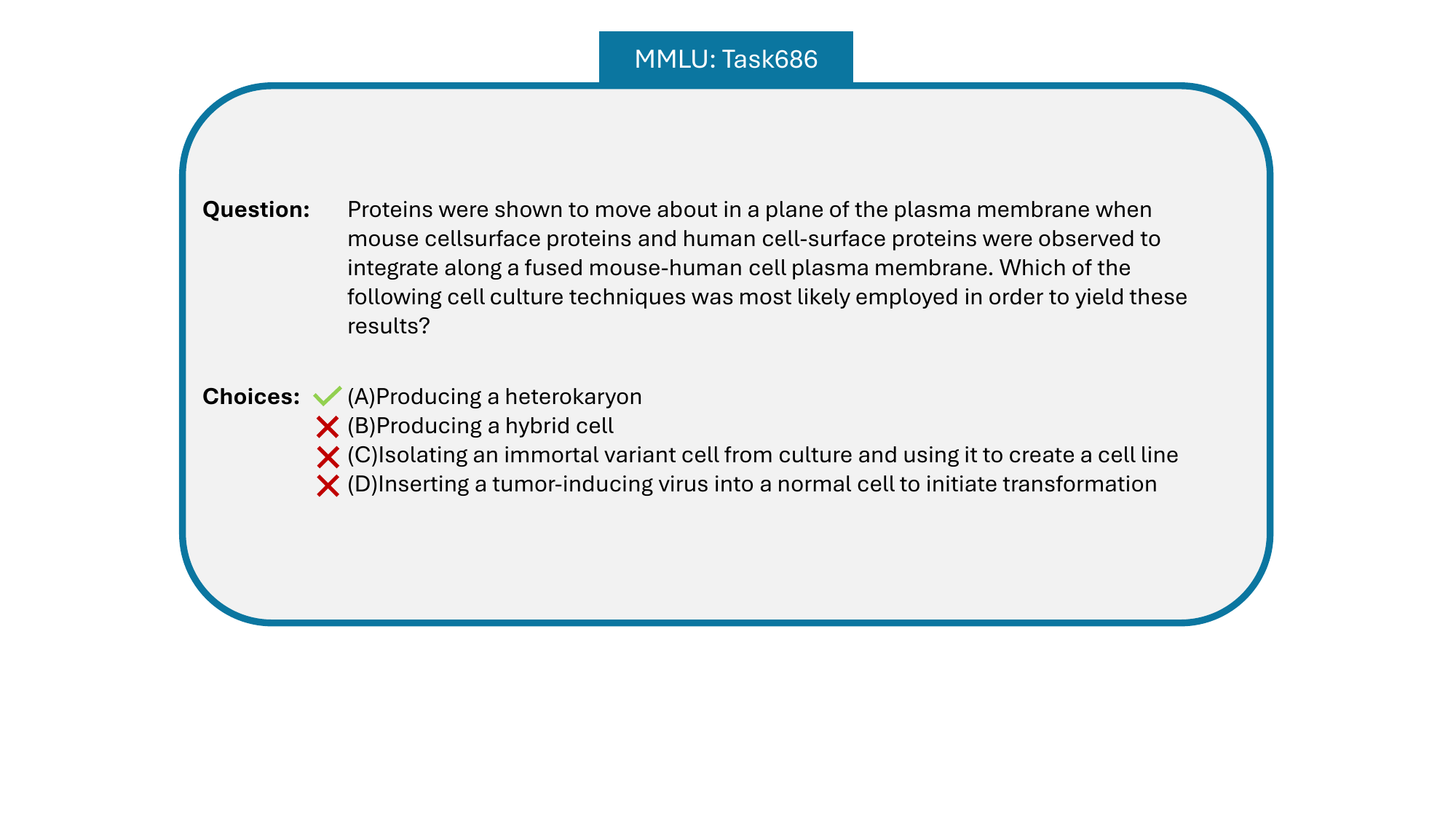}
    \caption{\textbf{Example for Task686 (MMLU).} An example from the MMLU dataset: "Task686". This question is from the biology domain, which requires domain knowledge at a college level.}
    \label{fig:task686}
\end{figure}
\begin{figure}[ht!]
    \centering
    \includegraphics[width=\textwidth]{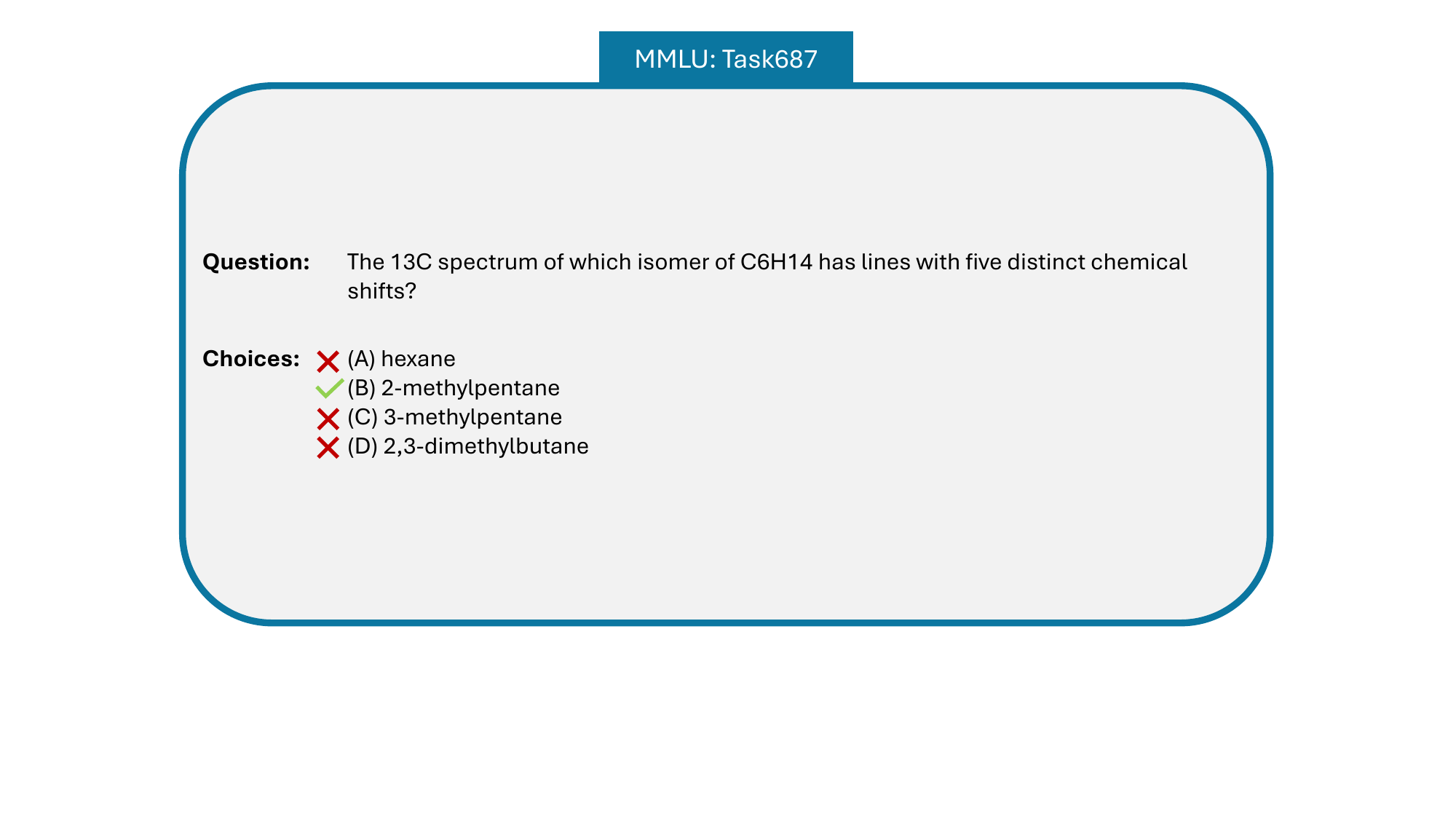}
    \caption{\textbf{Example for Task687 (MMLU).} An example from the MMLU dataset: "Task687". This question is from the chemistry domain, which requires domain knowledge at a college level.}
    \label{fig:task687}
\end{figure}
\begin{figure}[ht!]
    \centering
    \includegraphics[width=\textwidth]{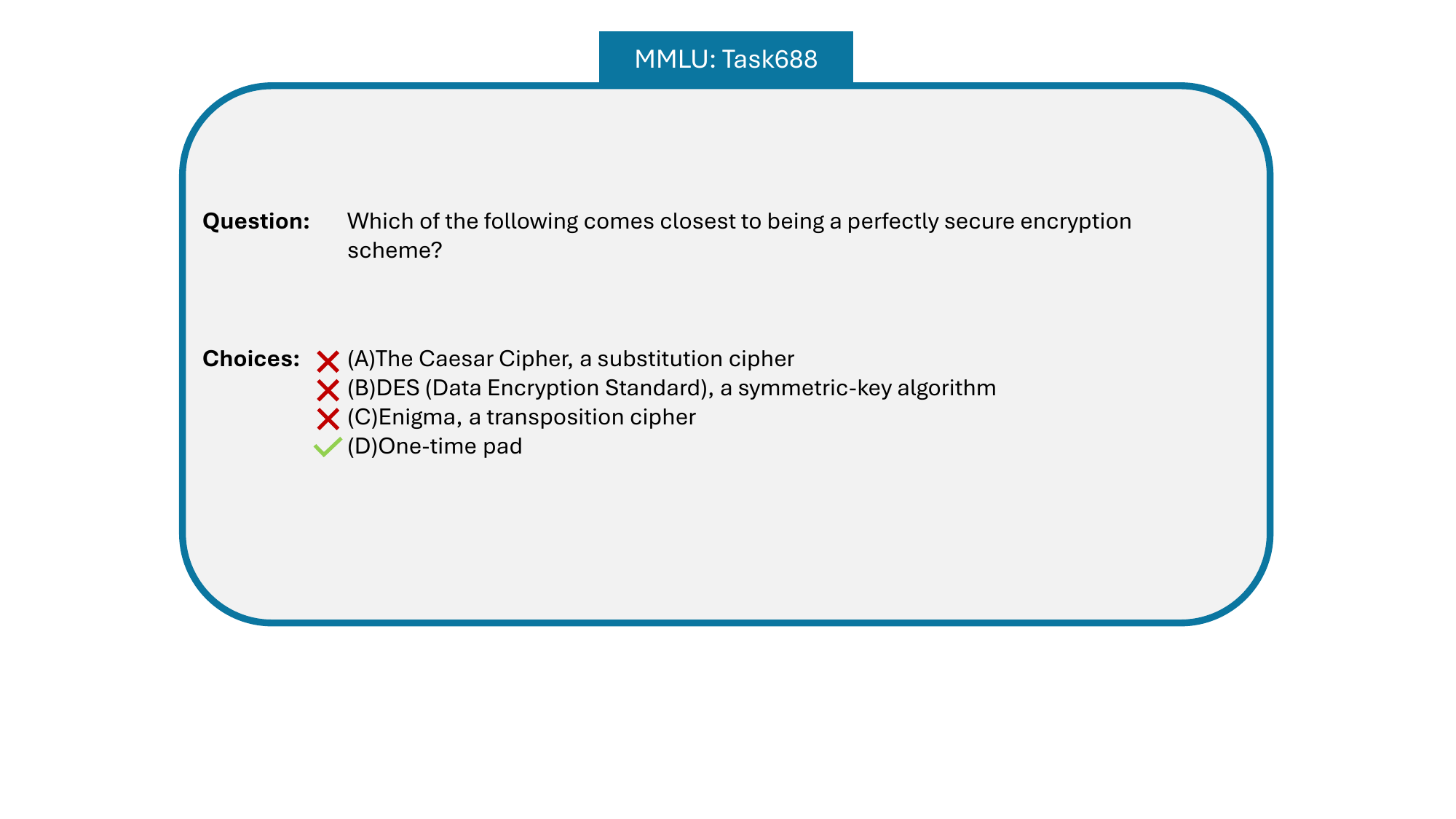}
    \caption{\textbf{Example for Task688 (MMLU).} An example from the MMLU dataset: "Task688". This question is from the computer science domain, which requires domain knowledge at a college level.}
    \label{fig:task688l}
\end{figure}
\begin{figure}[ht!]
    \centering
    \includegraphics[width=\textwidth]{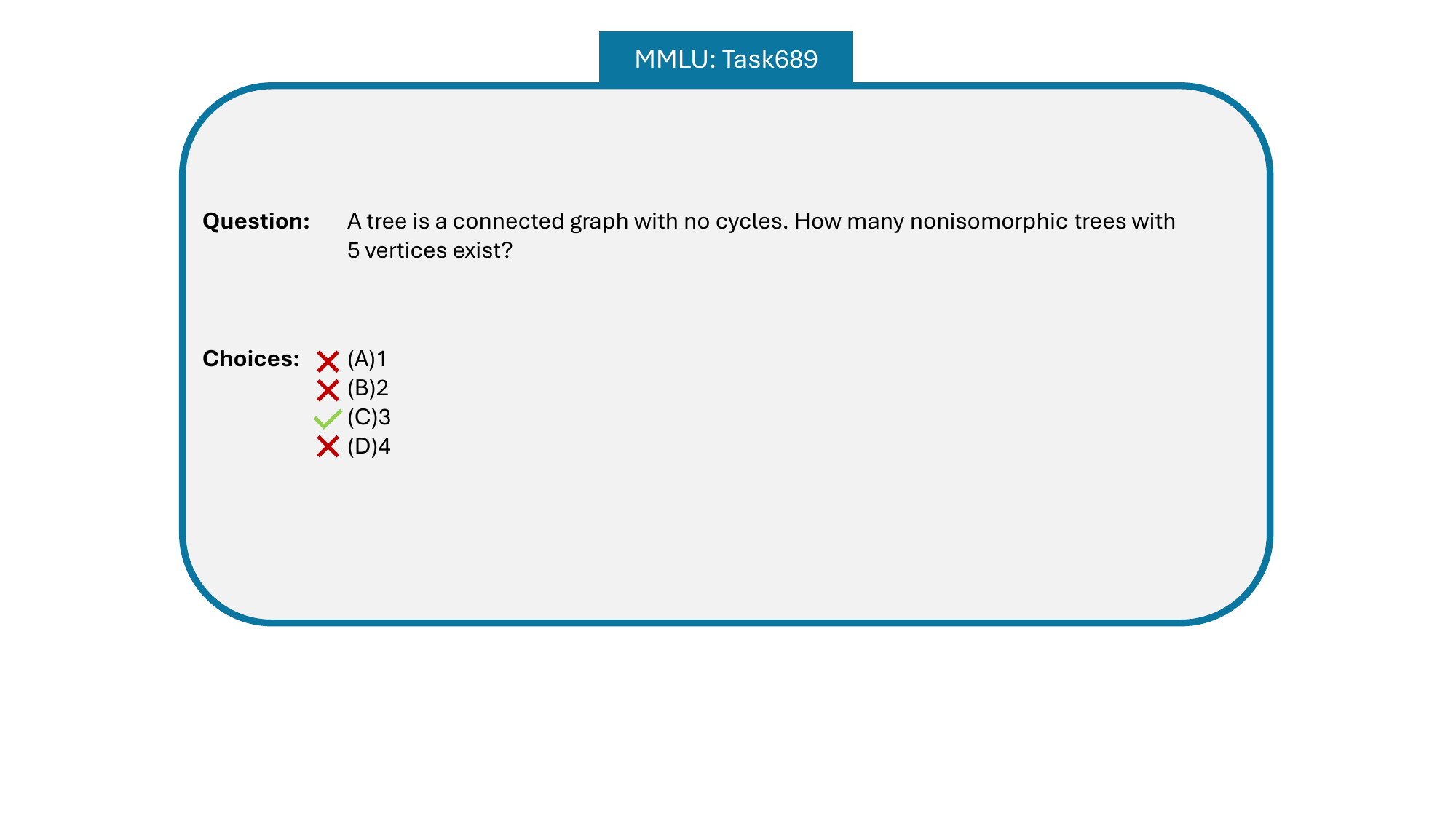}
    \caption{\textbf{Example for Task689 (MMLU).} An example from the MMLU dataset: "Task689". This question is from the mathematics domain, which requires domain knowledge at a college level.}
    \label{fig:task689}
\end{figure}
\begin{figure}[ht!]
    \centering
    \includegraphics[width=\textwidth]{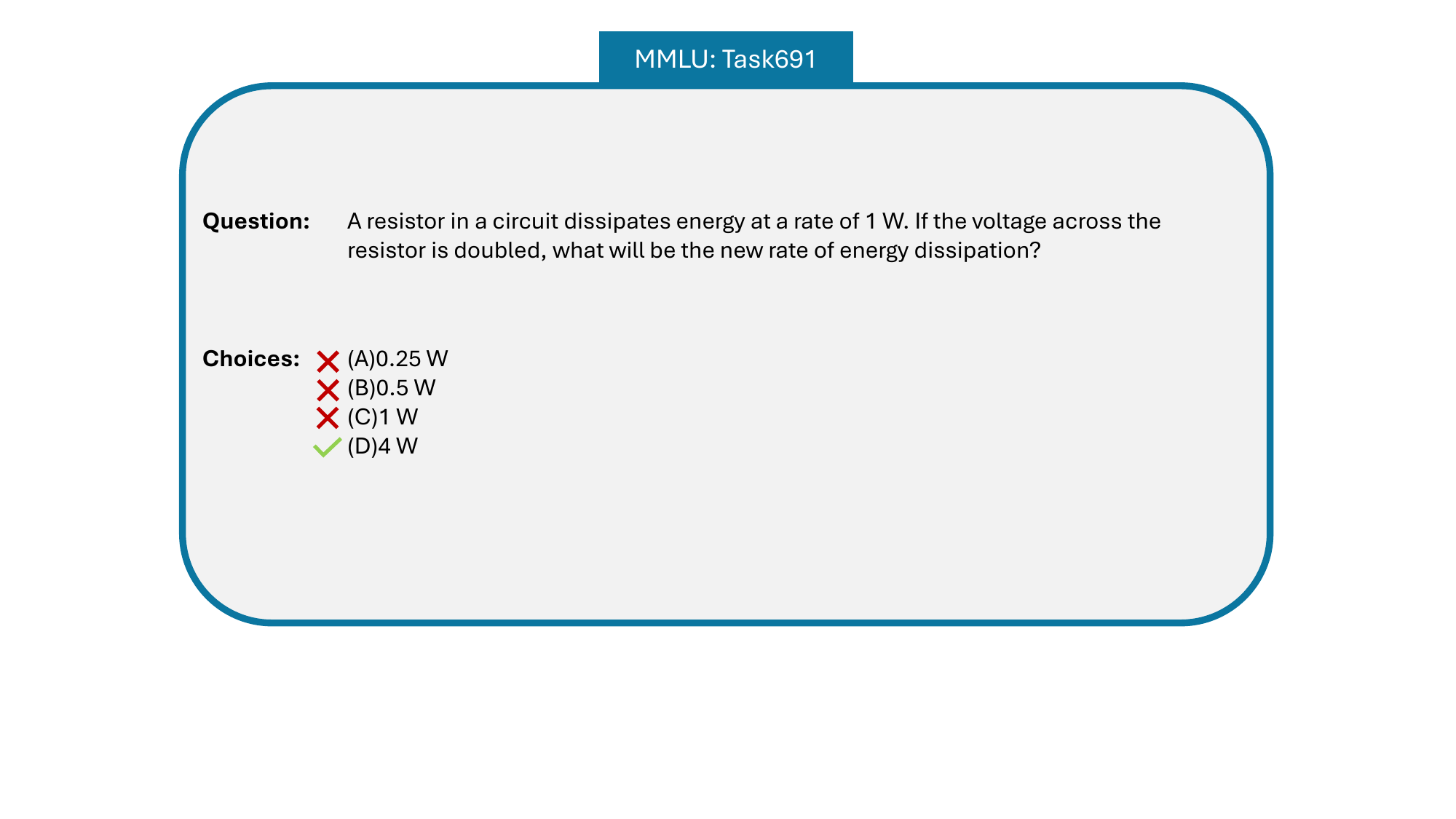}
    \caption{\textbf{Example for Task691 (MMLU).} An example from the MMLU dataset: "Task691". This question is from the physics domain, which requires domain knowledge at a college level.}
    \label{fig:task691}
\end{figure}
\begin{figure}[ht!]
    \centering
    \includegraphics[width=\textwidth]{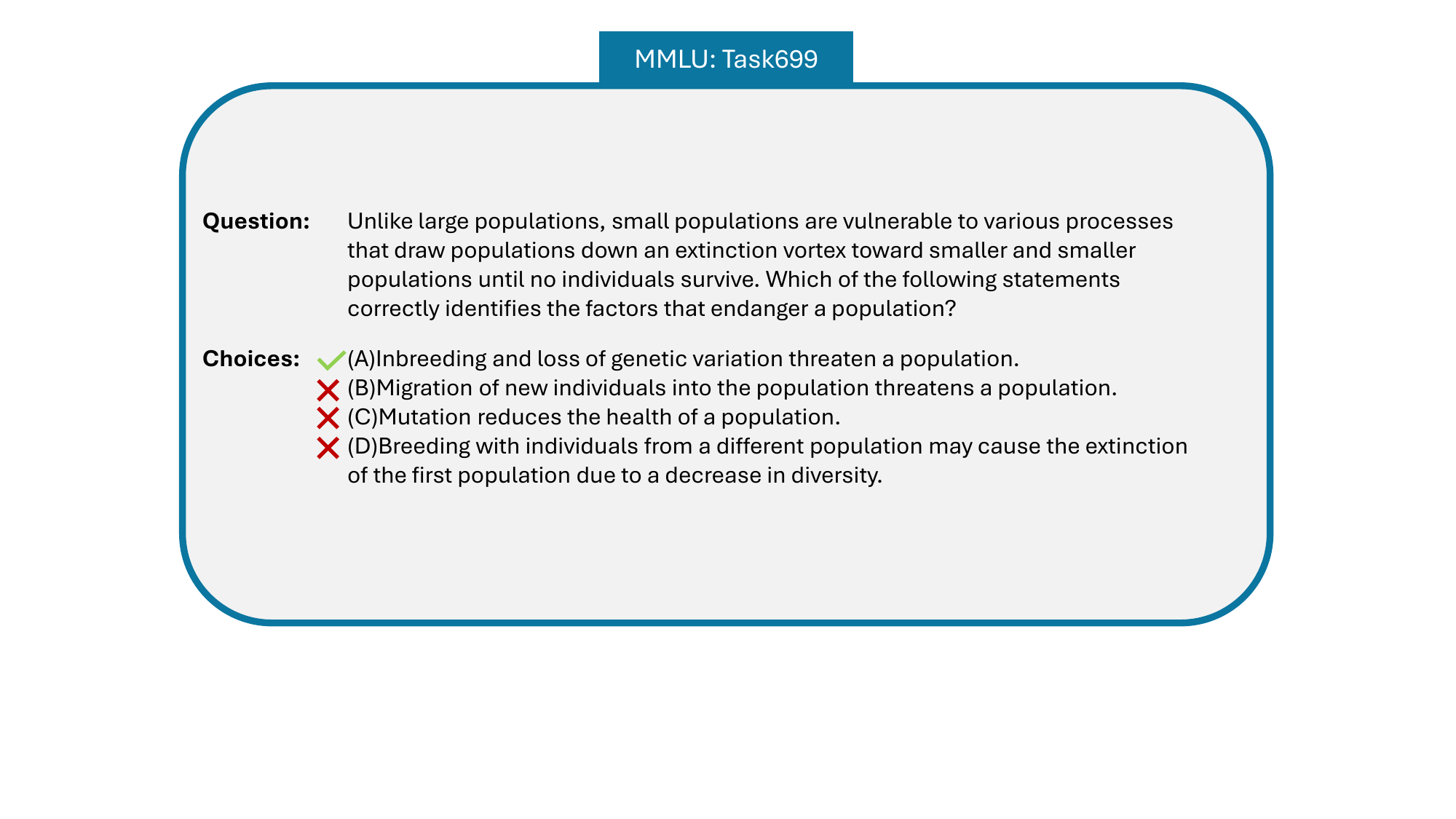}
    \caption{\textbf{Example for Task699 (MMLU).} An example from the MMLU dataset: "Task699". This question is from the biology domain, which requires domain knowledge at a high school level.}
    \label{fig:task699}
\end{figure}
\begin{figure}[ht!]
    \centering
    \includegraphics[width=\textwidth]{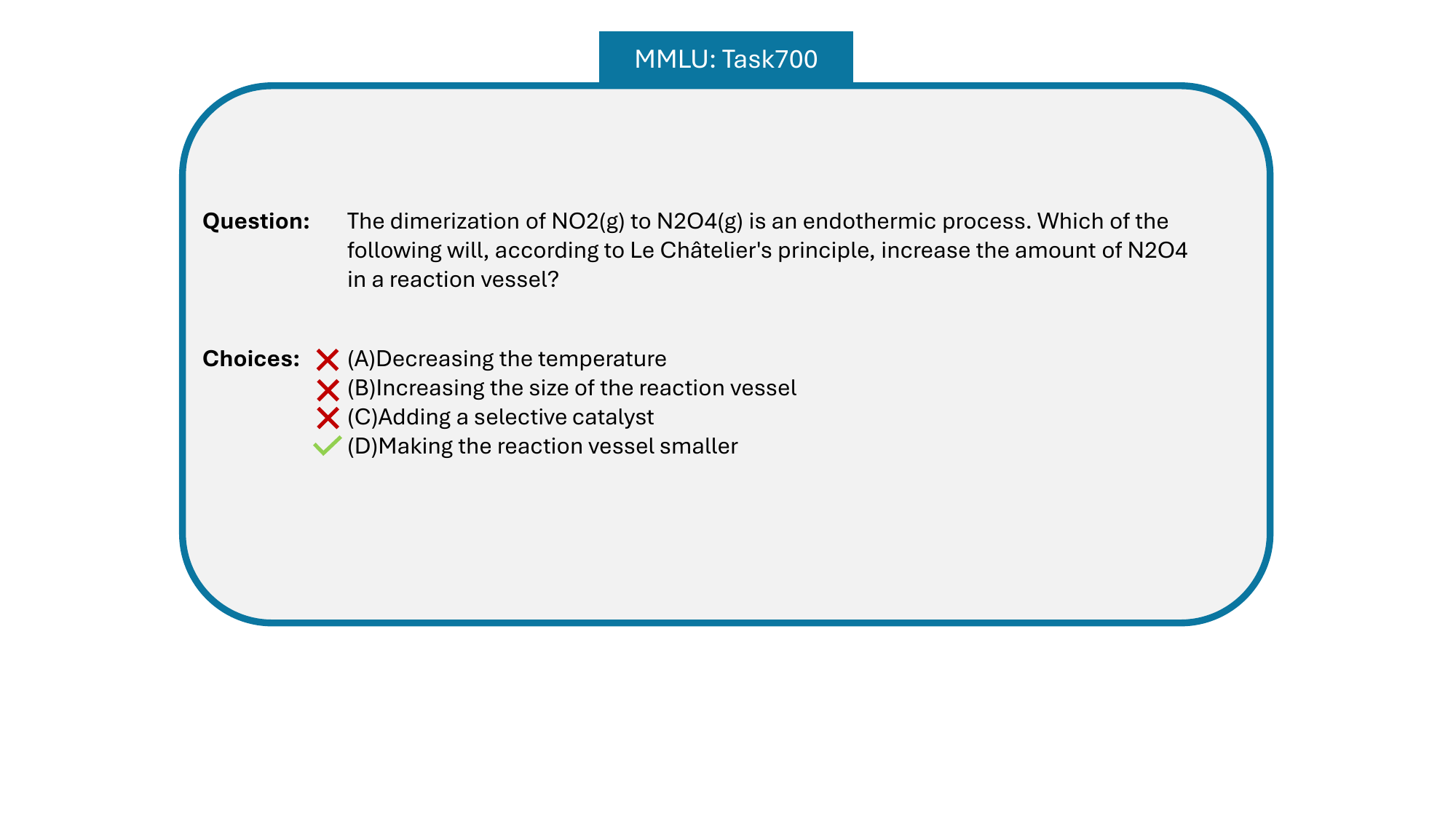}
    \caption{\textbf{Example for Task700 (MMLU).} An example from the MMLU dataset: "Task700". This question is from the chemistry domain, which requires domain knowledge at a high school level.}
    \label{fig:task700}
\end{figure}
\begin{figure}[ht!]
    \centering
    \includegraphics[width=\textwidth]{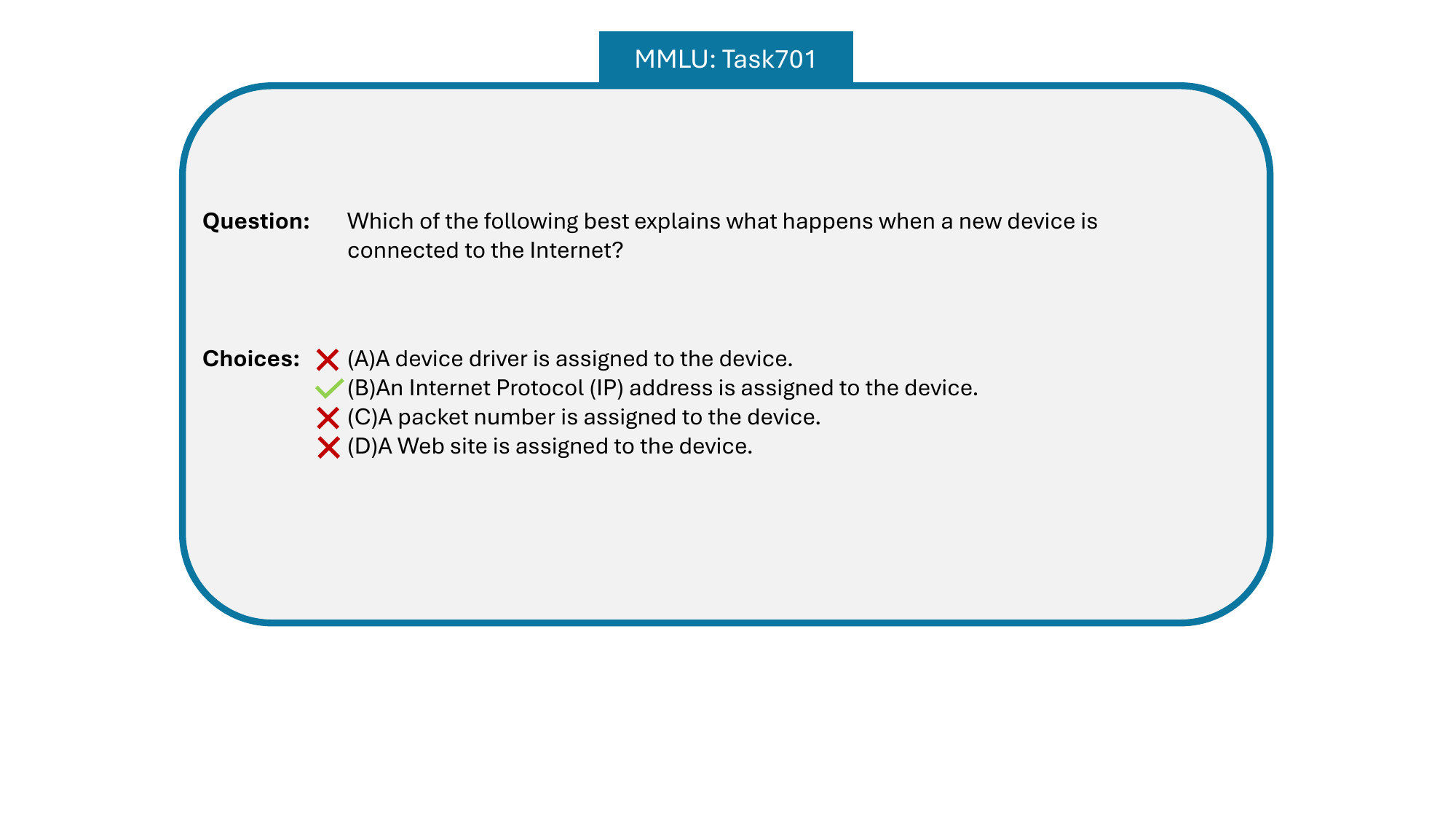}
    \caption{\textbf{Example for Task701 (MMLU).} An example from the MMLU dataset: "Task701". This question is from the computer science domain, which requires domain knowledge at a high school level.}
    \label{fig:task701}
\end{figure}
\begin{figure}[ht!]
    \centering
    \includegraphics[width=\textwidth]{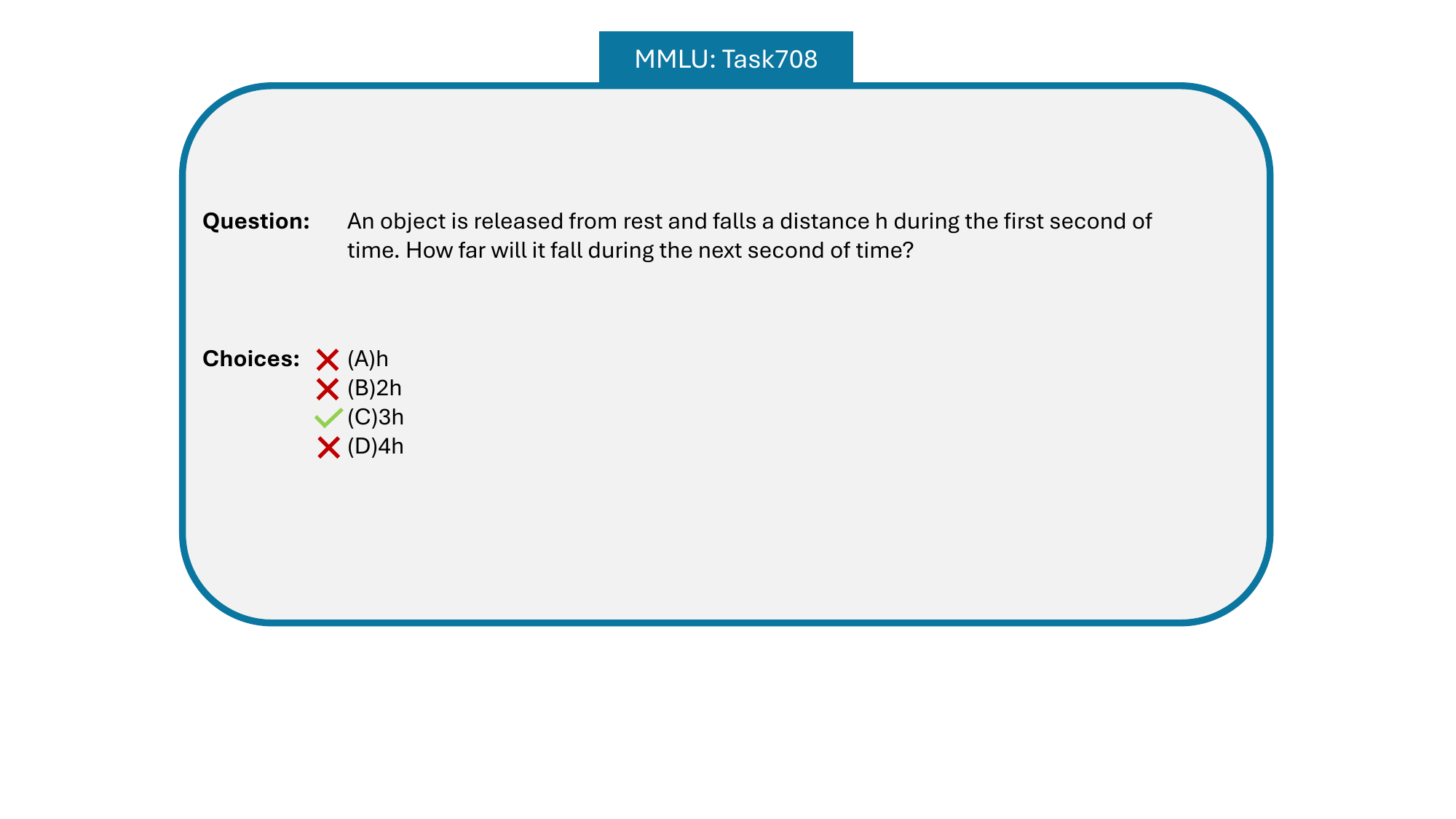}
    \caption{\textbf{Example for Task708 (MMLU).} An example from the MMLU dataset: "Task708". This question is from the physics domain, which requires domain knowledge at a high school level.}
    \label{fig:task708}
\end{figure}
\begin{figure}[ht!]
    \centering
    \includegraphics[width=\textwidth]{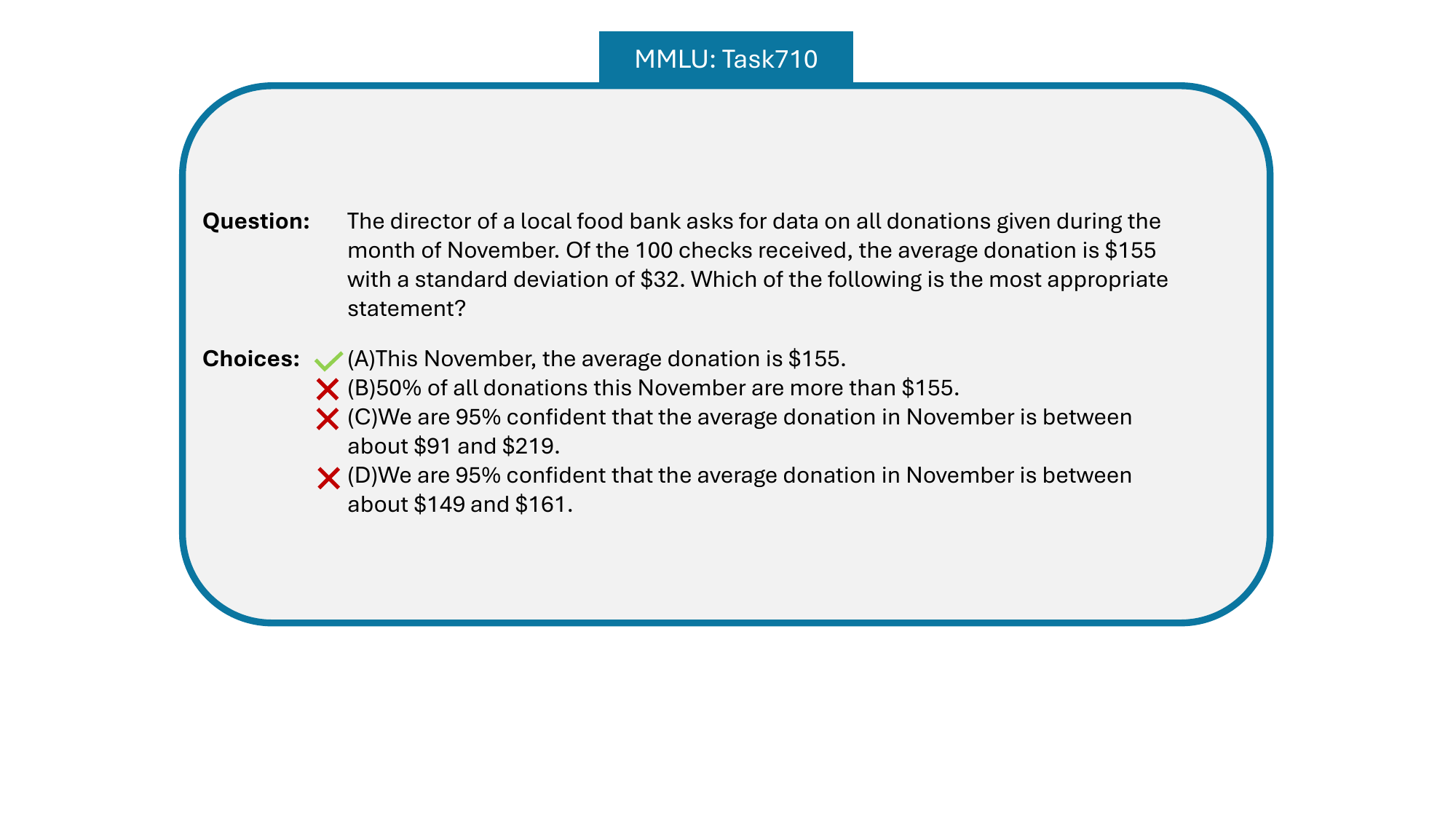}
    \caption{\textbf{Example for Task710 (MMLU).} An example from the MMLU dataset: "Task710". This question is from the statistics domain, which requires domain knowledge at a high school level.}
    \label{fig:task710}
\end{figure}
\begin{figure}[ht!]
    \centering
    \includegraphics[width=\textwidth]{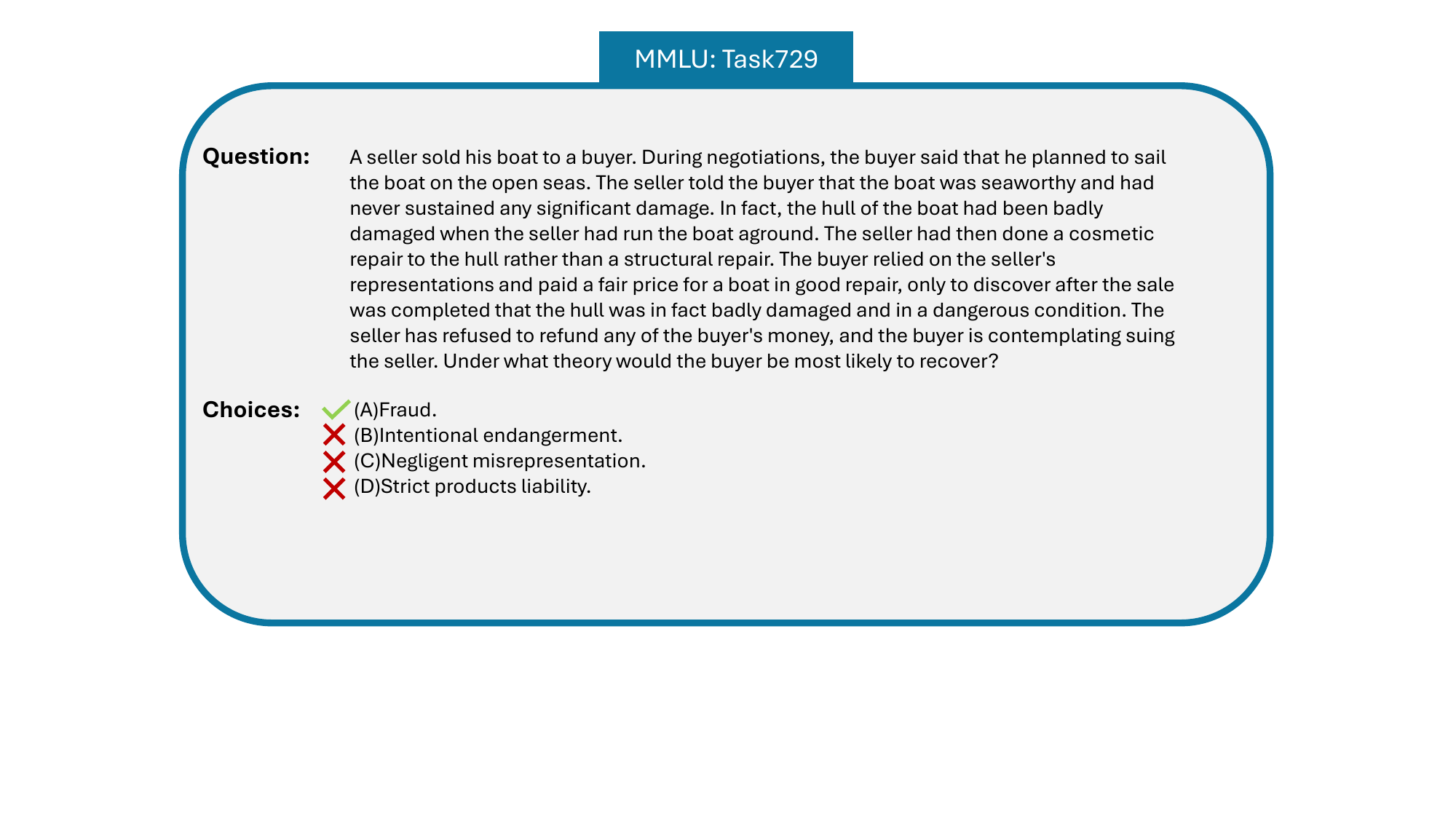}
    \caption{\textbf{Example for Task729 (MMLU).} An example from the MMLU dataset: "Task729". This question is from the law domain, which requires domain knowledge at a professional level.}
    \label{fig:task729}
\end{figure}
\begin{figure}[ht!]
    \centering
    \includegraphics[width=\textwidth]{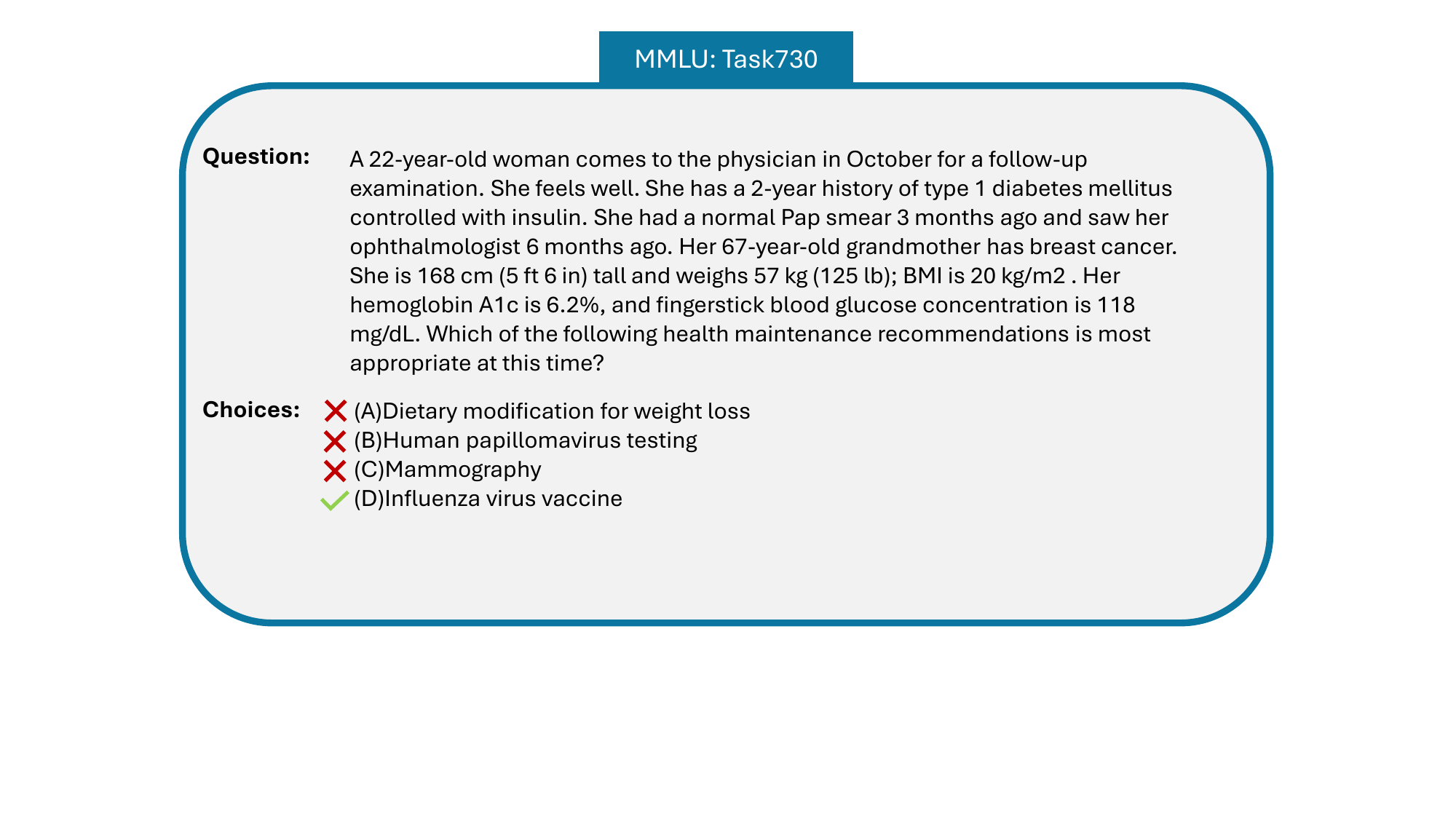}
    \caption{\textbf{Example for Task730 (MMLU).} An example from the MMLU dataset: "Task730". This question is from the medicine domain, which requires domain knowledge at a professional level.}
    \label{fig:task730}
\end{figure}
\begin{figure}[ht!]
    \centering
    \includegraphics[width=\textwidth]{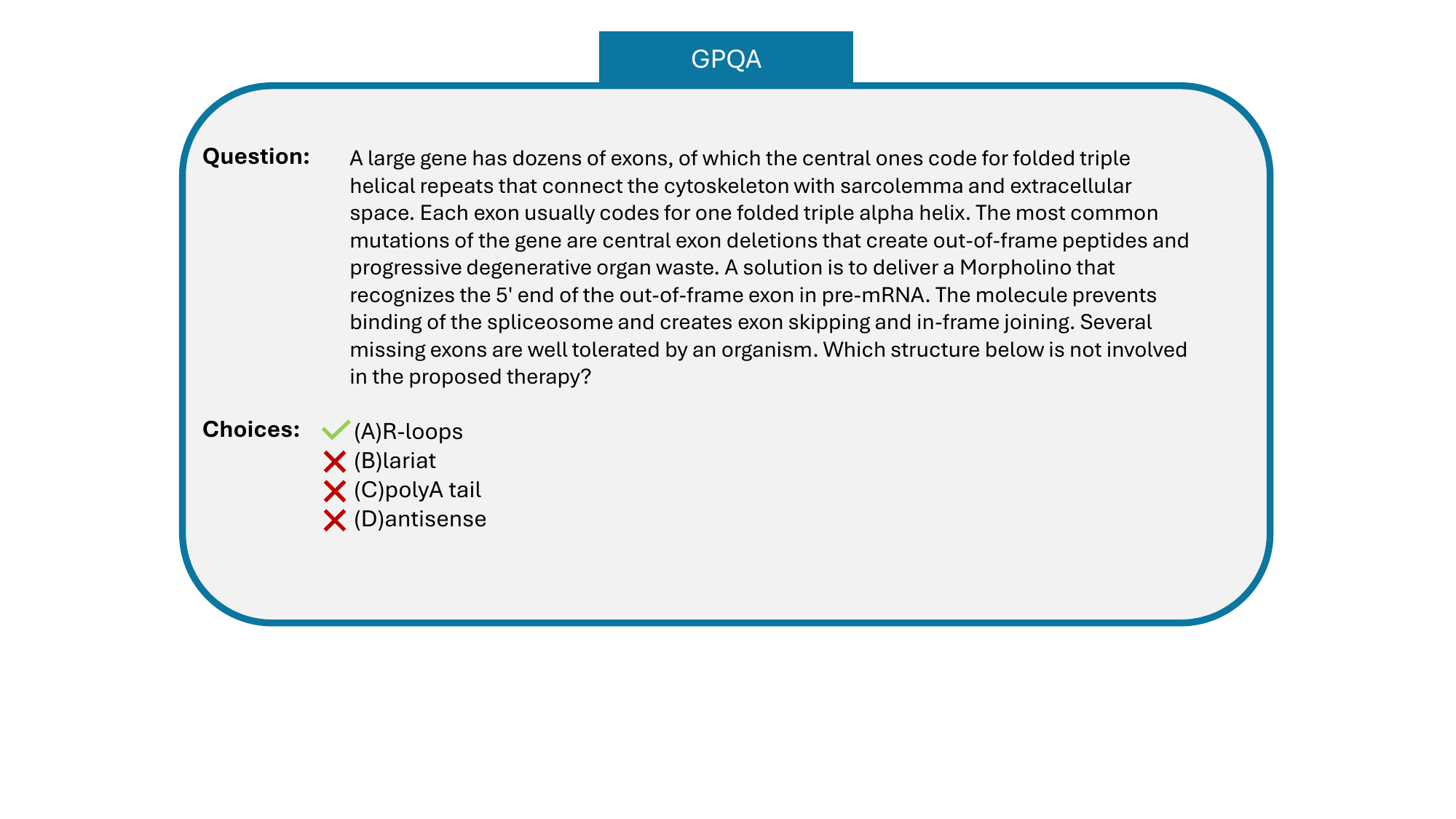}
    \caption{\textbf{Example for GPQA.} An example from the GPQA dataset. This question is from the biology domain, which requires domain knowledge at a PhD level.}
    \label{fig:gpqa}
\end{figure}\begin{figure}[ht!]
    \centering
    \includegraphics[width=\textwidth]{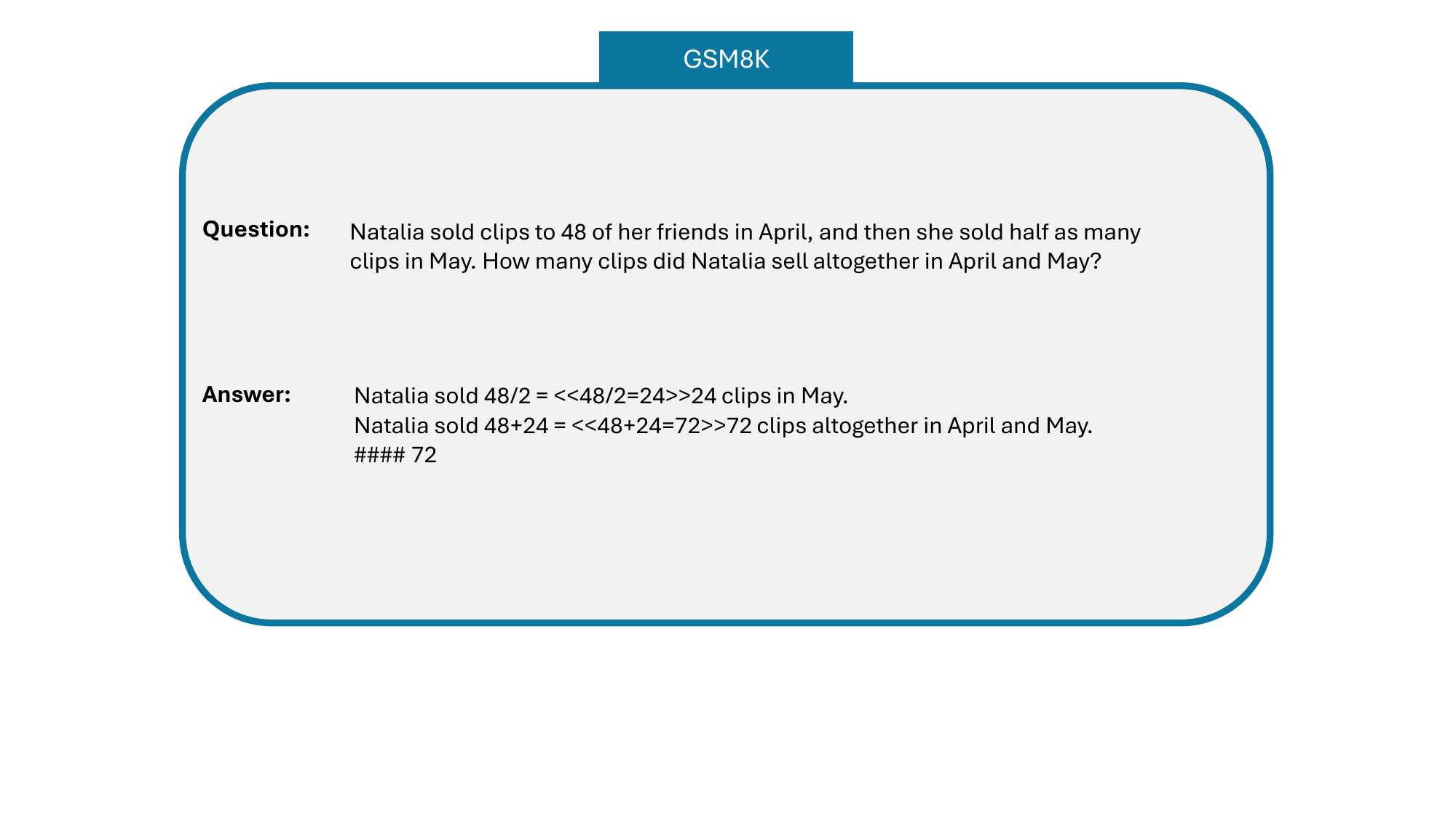}
    \caption{\textbf{Example for GSM8K.} An example from the GSM8K dataset. This question is from the mathematics domain, which requires domain knowledge at a grade school level.}
    \label{fig:gsm8kl}
\end{figure}
\setcellgapes{10pt} 
\makegapedcells

\renewcommand{\arraystretch}{2}
\definecolor{darkred}{rgb}{0.55, 0.0, 0.0}
\begin{table}[ht!]
    \centering 
    \adjustbox{max width=\textwidth}{
    \begin{tabular}{|c|c|c|}
    \hline
   Instruction & Specificity & Replacement Ratio\\
    \hline
    \hline
        \multirow{4}{*} { \makecell{\parbox{0.7\textwidth}{\justifying A breeder of \textcolor{darkred}{dogs} induced a \textcolor{darkred}{purchaser} to buy a puppy by representing that it was a registered basset \textcolor{darkred}{hound}, when in \textcolor{darkred}{fact} the breeder knew it was a mixed \textcolor{darkred}{breed}. The \textcolor{darkred}{purchaser} later discovered that the \textcolor{darkred}{representation} was false. She wants to sue to disaffirm the \textcolor{darkred}{contract} and get a refund. What legal \textcolor{darkred}{theory} would be best applicable to decide this \textcolor{darkred}{case}?}}}& \multirow{4}{*}{Original} & \multirow{4}{*}{-}\\
         & &\\
         & &\\ 
         & &\\\hline
         \multirow{4}{*} {\makecell{\parbox{0.7\textwidth}{\justifying A breeder of \textcolor{darkred}{domesticated\_animals} induced a \textcolor{darkred}{client} to buy a puppy by representing that it was a registered basset \textcolor{darkred}{hound\_dog}, when in \textcolor{darkred}{record} the breeder knew it was a mixed \textcolor{darkred}{variety}. The \textcolor{darkred}{client} later discovered that the \textcolor{darkred}{psychosexuality} was false. She wants to sue to disaffirm the \textcolor{darkred}{grant} and get a refund. What legal \textcolor{darkred}{explanation} would be best applicable to decide this \textcolor{darkred}{instance}?}}}& \multirow{4}{*}{Low} & \multirow{4}{*}{100\%}\\
         & &\\ 
         & &\\ 
         & &\\\hline
         \multirow{4}{*} {\makecell{\parbox{0.7\textwidth}{\justifying A breeder of \textcolor{darkred}{mongrels} induced a \textcolor{darkred}{emptor} to buy a puppy by representing that it was a registered basset \textcolor{darkred}{Afghan\_hound}, when in \textcolor{darkred}{basics} the breeder knew it was a mixed \textcolor{darkred}{animal\_group}. The \textcolor{darkred}{emptor} later discovered that the \textcolor{darkred}{version} was false. She wants to sue to disaffirm the \textcolor{darkred}{charter} and get a refund. What legal \textcolor{darkred}{atomistic\_theory} would be best applicable to decide this \textcolor{darkred}{bit}?}}}& \multirow{4}{*}{Intermediate} & \multirow{4}{*}{100\%}\\
         & &\\ 
         & &\\ 
         & &\\\hline
         \multirow{4}{*} {\makecell{\parbox{0.7\textwidth}{\justifying A breeder of \textcolor{darkred}{puppys} induced a \textcolor{darkred}{customer\_agent} to buy a puppy by representing that it was a registered basset \textcolor{darkred}{greyhound}, when in \textcolor{darkred}{rudiments} the breeder knew it was a mixed \textcolor{darkred}{bloodstock}. The \textcolor{darkred}{customer\_agent} later discovered that the \textcolor{darkred}{appearance} was false. She wants to sue to disaffirm the \textcolor{darkred}{adhesion\_contract} and get a refund. What legal \textcolor{darkred}{atomism} would be best applicable to decide this \textcolor{darkred}{humiliation}?}}}& \multirow{4}{*}{High} & \multirow{4}{*}{100\%}\\
         & &\\ 
         & &\\ 
         & &\\\hline
    \end{tabular}
    }
    \caption{\textbf{Examples for processed Instructions.} This is one example of a 100\% replacement of nouns for low, intermediate and high specificity. Words highlighted in \textcolor{darkred}{red} represent synonymizable nouns in the original sample, and in the modified samples, they denote the corresponding synonyms for low, intermediate, and high specificity.}
    \label{tab:examples}
\end{table}

\end{document}

%% file: statement.tex
\thispagestyle{empty}

\null
\vspace{16.5cm}

\rule{\textwidth}{0.4pt}

I hereby declare that I have written this thesis independently without any help from others and without the use of documents or aids other than those stated. I have mentioned all used sources and cited them correctly according to established academic citation rules.

\vspace{0.2cm}

Göttingen, \mysubmissiondate